\documentclass[journal]{IEEEtran}
\usepackage{lineno}
\usepackage{bbm}
\usepackage{bm}
\usepackage{amsfonts}
\usepackage{amsmath}
\usepackage{mathrsfs}
\usepackage{amssymb}
\usepackage{enumerate}
\usepackage{graphicx}
\usepackage{subfigure}
\usepackage{booktabs}
\usepackage{graphicx}
\usepackage{setspace}
\usepackage{algorithm}
\usepackage{algorithmic}
\usepackage{setspace}
\usepackage{multirow}
\usepackage{array}
\usepackage{color}
\usepackage{cite}
\usepackage{geometry}
\usepackage{hyperref}
\usepackage{bbding}

\ifCLASSINFOpdf
\else
\fi
\begin{document}
\title{Erasing, Transforming, and Noising Defense Network for Occluded Person Re-Identification}
\author{Neng~Dong, Liyan~Zhang, Shuanglin~Yan, Hao~Tang and Jinhui~Tang, \emph{Senior Member, IEEE}
\thanks{N. Dong, S. Yan, H. Tang, and J. Tang are with the School of Computer Science and Engineering, Nanjing University of Science and Technology, Nanjing 210094, China  (e-mail: neng.dong@njust.edu.cn; shuanglinyan@njust.edu.cn;
tanghao0918@njust.edu.cn;
jinhuitang@njust.edu.cn).}
\thanks{L. Zhang is with the College of Computer Science and Technology, Nanjing University of Aeronautics and Astronautics, Nanjing 210016, China (e-mail: zhangliyan@nuaa.edu.cn).}
}
\markboth{SUBMISSION OF IEEE Transactions on Circuits and Systems for Video Technology, 2023}%
{Shell \MakeLowercase{\textit{et al.}}: Bare Demo of IEEEtran.cls for IEEE Journals}
\maketitle
\begin{abstract}
Occlusion perturbation presents a significant challenge in person re-identification (re-ID), and existing methods that rely on external visual cues require additional computational resources and only consider the issue of missing information caused by occlusion. In this paper, we propose a simple yet effective framework, termed Erasing, Transforming, and Noising Defense Network (ETNDNet), which treats occlusion as a noise disturbance and solves occluded person re-ID from the perspective of adversarial defense. In the proposed ETNDNet, we introduce three strategies: Firstly, we randomly erase the feature map to create an adversarial representation with incomplete information, enabling adversarial learning of identity loss to protect the re-ID system from the disturbance of missing information. Secondly, we introduce random transformations to simulate the position misalignment caused by occlusion, training the extractor and classifier adversarially to learn robust representations immune to misaligned information. Thirdly, we perturb the feature map with random values to address noisy information introduced by obstacles and non-target pedestrians, and employ adversarial gaming in the re-ID system to enhance its resistance to occlusion noise. Without bells and whistles, ETNDNet has three key highlights: (i) it does not require any external modules with parameters, (ii) it effectively handles various issues caused by occlusion from obstacles and non-target pedestrians, and (iii) it designs the first GAN-based adversarial defense paradigm for occluded person re-ID. Extensive experiments on six public datasets fully demonstrate the effectiveness, superiority, and practicality of the proposed ETNDNet. The code will be released at \url{https://github.com/nengdong96/ETNDNet}.

\end{abstract}
\IEEEpeerreviewmaketitle
\section{Introduction}
\IEEEPARstart{P}{erson} re-identification (re-ID) aims at matching the same pedestrians from different cameras, providing technical support for a series of intelligent security tasks such as pedestrian tracking, autonomous driving, and activity recognition \cite{MCNF, DLAD, CCG-LSTM}. With the rapid development of the re-ID community, various superior algorithms have emerged \cite{BAGTRICKS, AGW, LSST, AADIFL, BPDA}. However, most of these methods ideally assume that the whole body of each pedestrian is visible, which is challenging to satisfy in real scenarios due to the inevitable occlusion. 
\begin{figure}[th!]
\centering
\subfigbottomskip=-1pt
\subfigcapskip=-1pt
\subfigure[] {\includegraphics[height=1.2in,width=3.3in,angle=0]{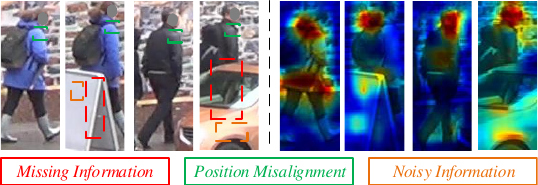}}
\subfigure[] {\includegraphics[height=1.0in,width=3.3in,angle=0]{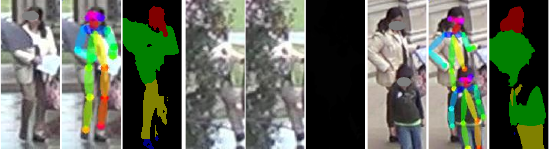}}
\caption{
(a) Challenges arising from occlusions, such as missing information, position misalignment, and noisy information, impact the re-ID model. Consequently, the re-ID model tends to prioritize local regions that are visible on occluded images while disregarding obscured and misaligned parts. Moreover, it may even allocate attention to noise areas associated with obstacles, leading to erroneous identification results.
(b) External tools may produce inaccurate visual cues in complex scenes. For instance, obstacles with similar material to clothing can be misidentified as the human body, and large obstacles can restrict the amount of information available, thus posing challenges in pedestrian detection. Moreover, external models may introduce noisy information from non-target pedestrians when they struggle to perform on images containing multiple pedestrians.
}
\label{Fig:1}
\vspace{-2mm}
\end{figure}

Empirically, occlusion mainly causes the following issues \cite{Survey}: i). \textbf{\emph{Missing Information}}. Obstacles such as warning signs can cause the loss of identity-related information, such as legs and even the torso. ii). \textbf{\emph{Position Misalignment}}. In case of occlusion, the detected pedestrian box only contains a part of the human body. Obviously, position misalignment occurs when holistic and partial images undergo the same scaling. iii). \textbf{\emph{Noisy Information}}. Besides the loss of valuable information, obstacles and non-target pedestrians introduce irrelevant noisy information. Affected by these issues, the model tends to focus only on locally visible regions in occluded images (neglecting obscured and misaligned parts) and may even pay attention to noise areas associated with obstacles (see Figure \ref{Fig:1}(a)). Existing methods \cite{PVPM, HOReID, RFC} often rely on external tools such as pose key-points \cite{Openpose} and human semantics \cite{LIP} to detect visible pedestrian regions for identity matching. While effective, this approach requires additional computation costs and only considers the issue of missing information. Moreover, the visual cues produced by external tools may be inaccurate in complex scenes (as shown in Figure \ref{Fig:1}(b)). Consequently, accurately locating the visible areas of the target pedestrian becomes challenging for the model, leading to errors in pedestrian identity matching.

Adversarial defense methods have gained significant attention due to their ability to protect Deep Neural Networks (DNNs) from noise perturbations \cite{Explaining, Stability, Gradient}. Considering that occlusion can be regarded as a kind of noise disturbance to re-ID models, it is plausible to protect re-ID models immune to occlusion perturbation from the perspective of adversarial defense. Previous research has attempted to solve the problem of occlusion from this perspective. For example, Huang \emph{et al}. \cite{AdverOccluded} generated adversarial occluded samples by substituting a region of the original images with all zeros, subsequently fine-tuning the re-ID model to enhance the robustness of learned features. Similarly, Zhao \emph{et al}. \cite{IGOAS} proposed an easy-to-hard erasing approach to generate aggressive occluded data, which encourages the network to pay attention to non-occluded regions. However, these methods rely on generating adversarial samples to endow re-ID models with the ability to defend against occlusion interference. On the one hand, this way of generating adversarial samples increases the training cost due to the expansion of the sample base. On the other hand, the generated data can only simulate the issue of missing information while ignoring position misalignment and noisy information. 
Additionally, adversarial samples obtained through perturbation on the original inputs are weakly aggressive, while strongly aggressive adversarial examples are crucial for training models with robust defense capabilities. Moreover, the aforementioned methods focus on the information in non-erased areas without enhancing its discriminability, which limits performance improvement.

According to the findings in \cite{Bugs}, the main culprit of the vulnerability of DNNs is the adversarial disturbance acting on sensitive features. Therefore, it is more effective to directly apply various perturbations to pedestrian feature maps, which not only eliminate the need for regenerative data but can also be more aggressive than adversarial samples since feature maps inherently contain rich semantic information. To distinguish from the above adversarial samples, we refer to such adversarial examples formed by perturbing feature maps as adversarial representations. In addition, drawing inspiration from Generative Adversarial Networks (GAN) \cite{GAN}, which use a generator to produce samples and a discriminator to distinguish between real and fake samples, we introduce a similar paradigm for occluded person re-ID. For the person re-ID task, the proposed paradigm mainly includes an extractor for learning pedestrian representations and a classifier for identification. While the classifier can accurately identify benign representations, it fails to classify disturbed representations. However, if the representations learned by the extractor are robust enough, even when affected by occlusion disturbances, the classifier can correctly identify the adversarial representations as it does for benign ones. Therefore, it is desirable to endow the re-ID system with the ability to defend against occlusion perturbation through an adversarial game between the feature extractor and identity classifier. Notably, our proposed approach, unlike the previous GAN-based adversarial defense paradigm~\cite{GAT}, is specifically tailored for person re-ID, an image retrieval task, and does not require learning to generate adversarial perturbations for original inputs.

Motivated by the above discussion, we propose an Erasing, Transforming, and Noising Defense Network (ETNDNet) to protect the re-ID system against occlusion perturbations. Specifically, to address the issue of missing information, we employ random erasing on the feature map to generate an adversarial representation, which maximizes the identity loss to optimize the classifier and simultaneously minimizes the loss to optimize the feature extractor. 
By continuously confronting each other, the feature extractor is driven to learn a robust representation that deceives the classifier into correctly identifying the attacked version, which indicates that the learned representation is immune to information loss. 
Moreover, we randomly select an area on the feature map and transform its pixels to another area, simulating position misalignment. Similarly, training the classifier enables it to identify the clean feature map accurately, but fails to classify the perturbed one. And in turn, training the feature extractor enables the classifier to assign the correct identity category to the adversarial representation, thereby ensuring that the feature map remains unaffected by misaligned information. 
Recognition performance is negatively impacted by noisy information from obstacles and non-target pedestrians, yet limited studies have attempted to overcome this challenge. To this end, we further apply random noise to the feature map and accordingly train the re-ID model in an adversarial way, enhancing the robustness of feature maps against noisy information. Importantly, the above paradigm does not require the introduction of redundant components, offering significant advantages in terms of model simplicity.

The main contributions of this paper can be summarized as follows:
\begin{itemize}
\item Without external auxiliary models and additional trainable modules, we develop a simple yet effective adversarial defense method to learn occlusion-robust pedestrian representations, in which the first GAN-based adversarial defense paradigm is designed for person re-ID tasks.

\item Our proposed ETNDNet endows the re-ID system with the ability to handle occlusion perturbations, alleviating the adverse effects of missing information and position misalignment on recognition performance. Particularly, our approach effectively addresses the problem of noisy information resulting from occlusions caused by obstacles and non-target pedestrians, which has often been overlooked in previous studies.

\item We conduct extensive experiments on six public datasets and compare the performance with existing state-of-the-art methods to demonstrate the superiority of the proposed method. Moreover, we evaluate the effectiveness and practicality of our ETNDNet through various aspects including ablation studies and analysis of model complexity.
\end{itemize}

The remainder of this paper is organized as follows: The related works are introduced in Section \uppercase\expandafter{\romannumeral2}; The details of the proposed method are elaborated in Section \uppercase\expandafter{\romannumeral3}; The experimental comparison and analysis are reflected in Section \uppercase\expandafter{\romannumeral4}; The conclusion of the present paper is summarized in Section \uppercase\expandafter{\romannumeral5}.

\section{Related Work}
\subsection{Holistic Person Re-ID}
Most existing re-ID studies assume that the pedestrian body is holistic and can be used to learn a complete representation. Hand-crafted feature-based methods \cite{SALF, LOMO, AIESS} adopt pre-designed descriptors to capture pedestrian characteristics. For example, Farenzena \emph{et al}. \cite{SALF} computed maximally stable color regions and recurrent high-structured patches to extract color information and texture details, respectively. Liao \emph{et al}. \cite{LOMO} proposed a local maximal occurrence algorithm that models salient representations to handle viewpoint and illumination changes. Moreover, Li \emph{et al}.\cite{AIESS} combined Gaussian of Gaussian (GOG) features with dictionary learning to reduce domain discrepancies across datasets. However, with the expansion of data, the applicability of such methods becomes unsatisfactory since they fail to exploit sample distribution adequately. To overcome these limitations, deep learning-based methods \cite{DML, PCB, WSS, MPLMV, SCFC} have emerged. For example, Yi \emph{et al}. \cite{DML} first introduced deep neural networks (DNNs) for person re-ID and built a siamese model to capture pedestrian similarities. Considering the importance of fine-grained information in re-ID, Sun \emph{et al}. \cite{PCB} developed a part-aware feature extraction method with an adaptive partitioning strategy to align body parts. Li \emph{et al}.\cite{TALMVR} exploited features from multiple camera views to enable multi-view imaginative reasoning. Recently, additional techniques such as BNNeck \cite{BAGTRICKS} and GeM \cite{AGW} have been developed, further advancing person re-ID tasks. Nevertheless, in crowded scenes where pedestrians are often occluded by obstacles or other persons, these methods struggle to achieve satisfactory performance. To address this, our work focuses on solving the occlusion problem in person re-ID while achieving excellent recognition rates in holistic scenarios.

\subsection{Occluded Person Re-ID}
Compared with holistic re-ID, occluded person re-ID is more practical and challenging, where the probe samples are occluded and the gallery database contains both holistic and occluded images. To encourage the model to focus on human body parts, Zhuo \emph{et al}. \cite{co-saliency} simulated occluded samples of holistic data using an occlusion simulator and introduced a co-saliency branch to assist the feature extractor in capturing identity-related information. Moreover, Kiran \emph{et al}. \cite{HG} adopted a teacher-student manner to match the distributions of intra- and inter-class distances of occluded and holistic data, thereby separating visible regions from occluded images. To extract discriminative features of visible regions, He \emph{et al}. \cite{FPR} inserted a foreground-aware pyramid reconstruction module that adaptively assigns weights to body and occlusion parts. Despite their effectiveness, learning occlusion-robust representations remains a significant challenge for these methods. Our proposed ETNDNet falls into the external model-free category, however, it endows the re-ID model with the intrinsic ability to defend against occlusion perturbations.

Recent studies have focused on guiding the re-ID model to extract features from the visible region by utilizing external tools. For example, Miao \emph{et al}.\cite{PGFA} integrated a pre-trained pose estimator into the re-ID model, generating pose-aware heatmaps that indicate if specific body parts are occluded. Similarly, Gao \emph{et al}. \cite{PVPM} developed a pose-guided attention mechanism and a visibility predictor to end-to-end match identities. To capture high-order relation information of human parts, Wang \emph{et al}. \cite{HOReID} treated local features learned by keypoints as graph nodes to model the topology structure of the human body. Leveraging the intrinsic connection between re-ID and human parsing, Zhang \emph{et al}. \cite{SORN} introduced a semantic branch to extract global and local features simultaneously. Moreover, Zheng \emph{et al}. \cite{PGFL-KD} proposed a pose-guided feature learning approach with knowledge distillation to obtain semantics-aligned representation. Hou \emph{et al}. \cite{RFC} performed person foreground segmentation and devised a spatial region feature completion module to reason about occluded areas. Nevertheless, the performance of these methods largely relies on the accuracy of pose landmarks and body parsing. Additionally, the use of the auxiliary model increases computational costs and may compromise the accuracy of visible cues. In contrast, our method overcomes these limitations and achieves state-of-the-art performance.

\subsection{Adversarial Attack and Defense}
The application of DNNs has led to significant advancements in various computer vision tasks \cite{BlockMix, Fewshot, KGS, Boosting}. However, Szegedy \emph{et al}. \cite{Intriguing} discovered that even highly performing models may produce incorrect inferences when imperceptible perturbations are applied to images. Building on this observation, Goodfellow \emph{et al}. \cite{Explaining} introduced adversarial examples by adding noise to original inputs using the fast gradient sign method. Kurakin \emph{et al}. \cite{Adversarial} proposed a multi-step attack method that iteratively classifies attacked samples into categories with the lowest classification probabilities. For person re-ID tasks, recent studies \cite{Advpattern, Transferable} have demonstrated that the re-ID system is also vulnerable to adversarial attacks, such as images of pedestrians wearing different clothes or captured by different cameras, which can lead to misidentification of individuals. Specifically, Wang \emph{et al}. \cite{Advpattern} designed the advPattern algorithm to explore the impact of clothing changes on recognition performance. To maintain visual quality in adversarial examples, Wang \emph{et al}. \cite{Transferable} employed differentiable multi-shot sampling to control the number of malicious pixels and proposed a novel perception loss to ensure inconspicuous attacks.

To mitigate the interference of noise to DNNs, the concept of adversarial defense has been introduced. Specifically, Zheng \emph{et al}. \cite{Stability} proposed joint fine-tuning of pre-trained models using clean samples and their corresponding distorted copies to improve the accuracy of predicting adversarial examples. Lyu \emph{et al}. \cite{Gradient} developed a family of gradient regularization techniques to constrain the optimization of DNNs and defend against perturbations. Gao \emph{et al}. \cite{DeepCloak} enhanced model robustness by removing irrelevant features through a mask layer positioned between the feature extractor and classifier. Building on these studies, Wang \emph{et al}. \cite{Multiexpert} proposed a multi-expert adversarial attack detection approach to differentiate perturbed examples for person re-ID tasks. Treating occlusion as a form of noise perturbation, Huang \emph{et al}. \cite{AdverOccluded} and Zhao \emph{et al}. \cite{IGOAS} drove models to extract robust features using brute-force adversarial defense that requires perturbed examples. In contrast, our method generates reasonable adversarial representations to simulate various occlusion perturbations and trains the feature extractor and identity classifier using a novel GAN-based adversarial defense paradigm, effectively defending the re-ID system against occlusion interference.

\begin{figure*}[th!]
  \centering
  \includegraphics[width=6.8in,height=3.9in]{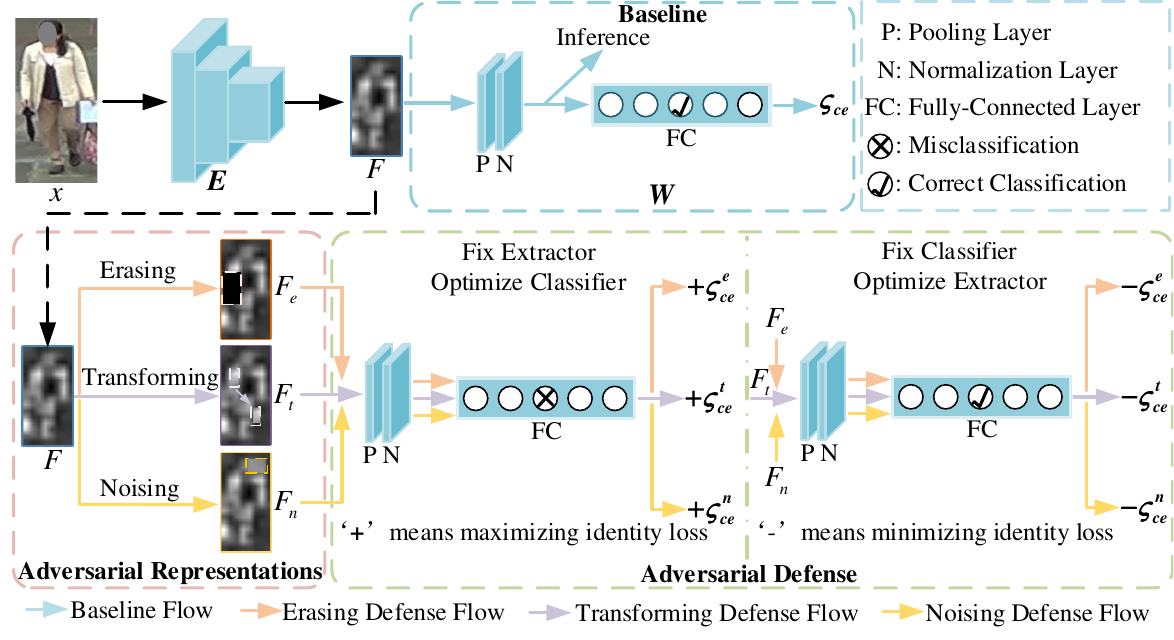}
  \vspace{-4mm}
  \caption{Overview of the proposed ETNDNet. Given an input pedestrian image, we employ a feature extractor to extract the corresponding feature map. This feature map is then forwarded to a classifier for identification purposes. To address the issue of recognition performance degradation due to missing information, we introduce a random erasing technique to generate an adversarial representation with incomplete information and iteratively optimize the extractor and classifier through a min-max game of identity loss to defend the re-ID model. Moreover, to mitigate the problem of position misalignment, we employ random pixel transformations within the feature map, enabling adversarial training of the feature extractor and identity classifier. Furthermore, we simulate the effect of occlusion by perturbing the feature map with random noise. This adversarial learning paradigm enhances the robustness of the feature map to noisy information. During inference, we only leverage the baseline branch exclusively to extract features from both the query and gallery images to compute similarity scores.}
  \label{Fig:2}
  \vspace{-2mm}
\end{figure*}

\section{The proposed methods} 

As illustrated in Figure \ref{Fig:2}, the proposed ETNDNet consists of a baseline framework and an occlusion defense system. The former is a standard re-ID model, which is trained to ensure the discriminativeness of learned features. The latter includes erasing defense, transforming defense, and noising defense modules. In these modules, we perturb the feature maps in different manners to generate multiple adversarial representations that can simulate various issues caused by occlusion, and accordingly train the feature extractor and identity classifier in an adversarial way to endow the re-ID model with the ability to defend against occlusion perturbation, thereby promoting the occlusion robustness of the learned features. In the following, we will describe the key technologies of each part in detail. 

\subsection{Baseline Framework}

We adopt the standard re-ID model as our baseline framework, which is commonly employed for the occluded person re-ID task. This model contains a feature extractor with ResNet-50 \cite{Resnet} pre-trained on ImageNet \cite{Imagenet} and an identity classifier composed of a pooling layer, a normalization layer, and a fully-connected layer. As illustrated in Figure \ref{Fig:2}, given a pedestrian image $x$, the feature extractor is responsible for learning representation, and the classifier is utilized to predict which identity the person belongs to. Let $\boldsymbol E$ and $\boldsymbol W$  respectively represent the feature extractor and identity classifier, we deploy the cross-entropy loss to optimize them:
\begin{equation}
\mathcal{L}_{ce}( \boldsymbol E, \boldsymbol W)=\sum_{i=1}^{N}-q_{i}log(p_{i}),
\end{equation}
where $N$ is the number of pedestrians, $p_{i}$ denotes the prediction logits of class $i$, $q_{i}$ is associated with the ground truth $y$:
\begin{equation}
q_{i}=\left\{
\begin{aligned}
&1- \frac{N-1}{N}\epsilon,&y=i \\
&\frac{\epsilon}{N},&y\ne i \\
\end{aligned}
~,\right.
\end{equation}
where $\epsilon$ is a constant and set to be 0.1. The association is derived from the label smoothing (LS) \cite{LS} technique, which effectively avoids the risk of overfitting.

\subsection{Adversarial Representations}
The above baseline branch ensures the basic discriminability of learned features. However, the model is vulnerable to occlusion interference. As mentioned before, it is feasible to train an occlusion-robust network from the perspective of adversarial defense. In this process, generating reasonable adversarial examples to simulate occlusion issues is crucial. The adversarial samples generated in \cite{AdverOccluded, IGOAS} can achieve this purpose. However, they are less aggressive and can only simulate the issue of missing information. To this end, we apply perturbation on pedestrian feature maps by erasing, transforming, and noising to generate multiple adversarial representations to simulate various occlusion issues.

\textbf{Erasing}. We randomly erase the feature map to acquire an adversarial representation with incomplete information, which can simulate the issue of pedestrian identity information loss caused by occlusion. Suppose the feature map is $F\in \mathbb{R}^{H\times W\times C}$, where $H$, $W$, and $C$ denote the height, width, and the number of channels, respectively. Similar to \cite{RE}, we first compute the spatial size $S=H\times W$ of the feature map. Next, we randomly initialize a proportion $\alpha_{e}$ to determine the size $S_{e}=\alpha_{e}\times S$ of the erasing rectangle region $R_{e}$, where $\alpha_{e} \in [0.02, 0.4]$. And then, we arbitrarily take an aspect ration $r_{e}$ and accordingly compute the height $H_{e}=\sqrt{S_{e}\times r_{e}}$ and width $W_{e}=\sqrt{S_{e}/r_{e}}$ of $R_{e}$, where $r_{e}\in [0.3, 1/0.3]$. Finally, we randomly select a point $(x_{e},y_{e})$ in $F$. If $x_{e} + W_{e} \leq W$ and $y_{e} + H_{e} \leq H$, $R_{e}=(x_{e}:x_{e}+W_{e},y_{e}:y_{e}+H_{e})$ can be determined. Otherwise, the above process is repeated until an appropriate $R_{e}$ is selected. Let each pixel belonging to $R_{e}$ be 0, and the adversarial representation $F_{e}$ can be formulated as:
\begin{equation}
F_{e}^{h,w}=\left\{
\begin{aligned}
&0,&{if}~{h,w}\in R_{e} \\
&F^{h,w},&otherwise \\
\end{aligned}
~.\right.
\end{equation}

\textbf{Transforming}. Occlusion may cause the pedestrian information at a certain position on the non-occluded image to appear at another position on the occluded image. Therefore, we randomly transform the feature map to simulate such a position misalignment issue, which is an ability that the adversarial samples do not possess. Firstly, we randomly initialize a proportion $\alpha_{t}$ and an aspect ratio $r_{t}$ to determine the height $H_{t}=\sqrt{\alpha_{t}HW\times r_{t}}$ and width $W_{t}=\sqrt{\alpha_{t}HW/r_{t}}$ of areas to be transformed, where $\alpha_{t} \in [0.02, 0.4]$ and $r_{t}\in [0.3, 1/0.3]$. Next, arbitrarily choosing two points $(x_{t_{1}},y_{t_{1}})$, and $(x_{t_{2}},y_{t_{2}})$, if $x_{t_{1}} + W_{t} \leq W$, $y_{t_{1}} + H_{t} \leq H$, $x_{t_{2}} + W_{t} \leq W$, and $y_{t_{2}} + H_{t} \leq H$, two areas $R_{t_{1}}=(x_{t_{1}}:x_{t_{1}}+W_{t},y_{t_{1}}:y_{t_{1}}+H_{t})$ and $R_{t_{2}}=(x_{t_{2}}:x_{t_{2}}+W_{t},y_{t_{2}}:y_{t_{2}}+H_{t})$ can be located. Accordingly, we transform the pixels of $R_{t_{1}}$ to $R_{t_{2}}$ for generating an adversarial representation with misaligned information $F_{t}$.

\textbf{Noising}. The feature map extracted from an image occluded by obstacles and non-target pedestrians often contains some noise irrelevant to the target pedestrian, which is detrimental to the re-ID system for identification. To this end, we further perturb the feature map by noising to generate an adversarial representation with noisy information. Specifically, with the initialized $\alpha_{n}$, $r_{n}$, and $(x_{n}, y_{n})$, we take a noising region $R_{n}=(x_{n}:x_{n}+W_{n},y_{n}:y_{n}+H_{n})$ and replace its pixels with random values that follow a uniform distribution. The value range of $\alpha_{n}$ and $r_{n}$, and the calculation way of $W_{n}$ and $H_{n}$ are the same as the above erasing process. Let $F_{n}$ denote the adversarial representation, it satisfies the following formula:
\begin{equation}
F_{n}^{h,w}=\left\{
\begin{aligned}
&\texttt{random.uniform}(0,1),&{if}~{h,w}\in R_{n} \\
&F^{h,w},&otherwise \\
\end{aligned}
~.\right.
\end{equation}

Note that the erased, transformed, and noised regions are the same for all images in a mini-batch, which prevents the model from being hard to converge \cite{IGOAS}.

\subsection{Adversarial Defense}

Existing adversarial defense studies on person re-ID mainly focus on training the model jointly with adversarial samples and original inputs, which is time-consuming and poorly defensive. Inspired by the adversarial training in GAN \cite{GAN}, which optimizes a generator and a discriminator in an adversarial way to encourage the former to generate fake samples that make the latter can not distinguish them from the real ones. For person re-ID, we expect the feature extractor to learn a sufficiently robust representation whose perturbed version can be correctly identified by the classifier as well. To this end, with the above adversarial representations, we develop three modules with a novel GAN-based adversarial defense strategy to overcome the issues of missing information, position misalignment, and noisy information caused by occlusion. No additional networks are required, each module employs adversarial training between the feature extractor and identity classifier, simple yet effective. 

\textbf{Erasing Defense}. Compared to the clean feature map $F$ that can be correctly identified (Baseline Branch), the adversarial representation $F_{e}$ loses some identity-related cues due to the information of $R_{e}$ being erased, leading the classifier into misidentification. Therefore, we maximize the cross-entropy loss to optimize identity classifier $\boldsymbol W$:
\begin{equation}
\mathop{\arg\max}\mathcal{L}_{ce}^{e}(\boldsymbol W)=\sum_{i=1}^{N}-q_{i}log(p^{e}_{i}),
\end{equation}
where $p_{i}^{e}$ is the probability that $F_{e}$ is recognized as category $y$. 

For the feature extractor $\boldsymbol E$, we expect the learned feature map $F$ from it can well depict the pedestrian characteristics, even if part of the information is lost, it can be correctly identified. To achieve this goal, we utilize $F_{e}$ to update the parameters of $\boldsymbol E$ by minimizing the cross-entropy loss:
\begin{equation}
\mathop{\arg\min}\mathcal{L}_{ce}^{e}(\boldsymbol E)=\sum_{i=1}^{N}-q_{i}log(p^{e}_{i}).
\end{equation}

\textbf{Transforming Defense}. Given the adversarial representation $F_{t}$, since its information at $R_{t_{2}}$ is inconsistent with the information at $R_{t_{2}}$ on the feature map $F$, the classifier fails to accurately identify the target pedestrian, which can be formulated as:
\begin{equation}
\mathop{\arg\max}\mathcal{L}_{ce}^{t}(\boldsymbol W)=\sum_{i=1}^{N}-q_{i}log(p^{t}_{i}),
\end{equation}
where $p_{i}^{t}$ is the prediction result of $F_{t}$ belonging to the class $i$.

Conversely, if the feature map $F$ enables the classifier to correctly identify its attacked version $F_{t}$, it means that $F$ is not affected by the misaligned information. To this end, we constrain $\boldsymbol E$ to satisfy the following formula:
\begin{equation}
\mathop{\arg\min}\mathcal{L}_{ce}^{t}(\boldsymbol E)=\sum_{i=1}^{N}-q_{i}log(p^{t}_{i}).
\end{equation}

\textbf{Noising Defense}. In order to further endow $F$ with the ability to resist the interference of noisy information, similar to the above processes, we utilize the perturbed representation $F_{n}$ to maximize the cross-entropy loss for optimizing $\boldsymbol W$, and in turn, minimizing the cross-entropy loss for optimizing $\boldsymbol E$. Suppose that the probability of $F_{n}$ being correctly classified is $p^{n}_{i}$, the above optimization strategy can be expressed as follows:
\begin{equation}
\mathop{\arg\max}\mathcal{L}_{ce}^{n}(\boldsymbol W)=\sum_{i=1}^{N}-q_{i}log(p^{n}_{i}),
\end{equation}
\begin{equation}
\mathop{\arg\min}\mathcal{L}_{ce}^{n}(\boldsymbol E)=\sum_{i=1}^{N}-q_{i}log(p^{n}_{i}).
\end{equation}

\begin{figure}[t!]
  \centering
  \includegraphics[width=0.9\linewidth]{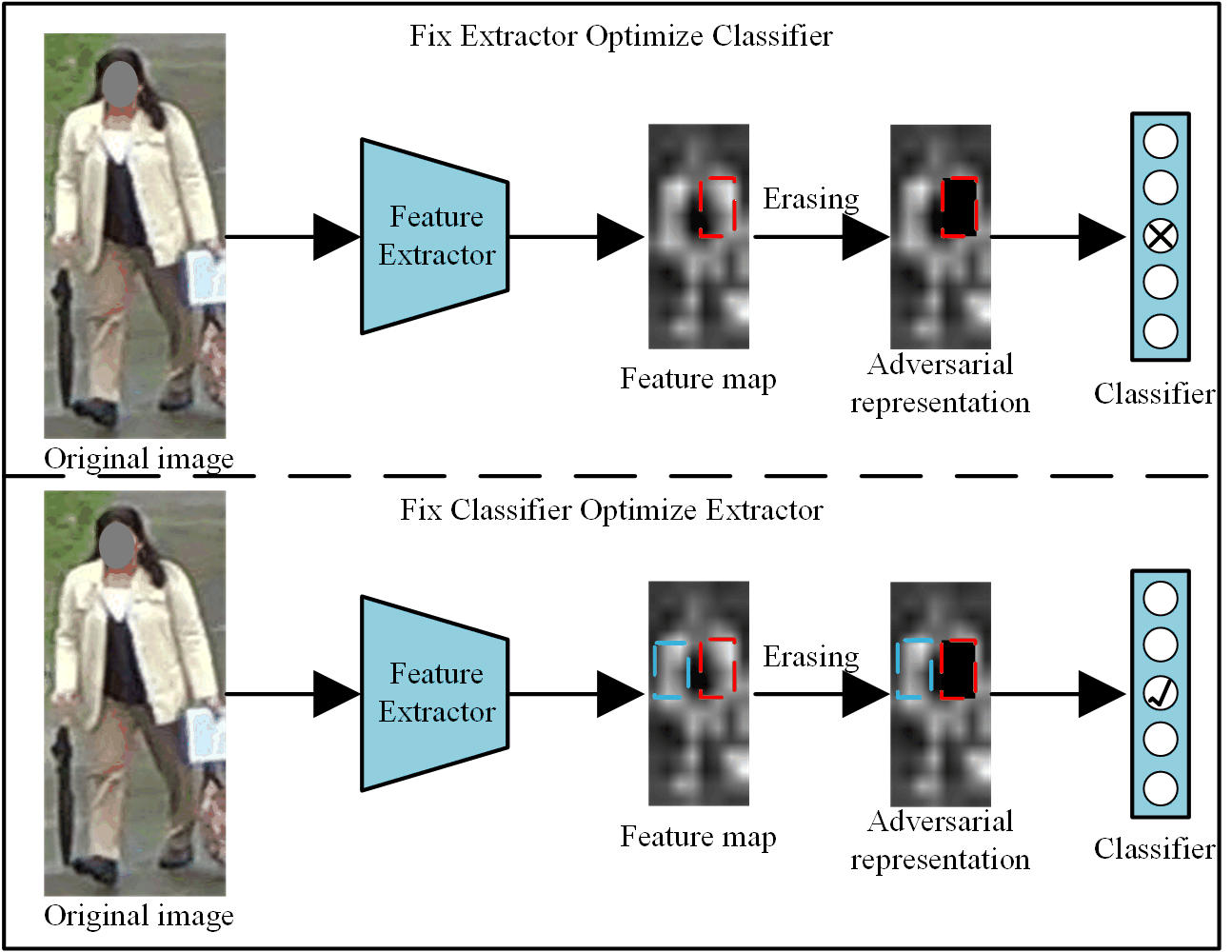}\\
  \caption{Erasing defense module. Through an adversarial game between the identity classifier and feature extractor, the model focuses on non-erased areas with highly discriminative information (the part marked by the blue box).}
  \label{Fig:3}
\end{figure}

Note that in all the above adversarial learning processes, the feature extractor is fixed when training the identity classifier and vice versa. Through continuous gaming and optimization between them, the re-ID system is able to defend against the perturbation of various occlusion issues. To gain a better understanding of the above adversarial game process, we take the erasing defense module as an example to further illustrate. As shown in Figure \ref{Fig:3}, assuming that the erased area on the feature map is the part marked by the red box. Initially, we fix the feature extractor and optimize the classifier by maximizing the identity loss. This enhances the discriminative capability of the classifier’s class prototype and improves its correct classification threshold. Next, we fix the classifier and optimize the feature extractor by minimizing the identity loss, which ensures that even if the information is erased, the feature map extracted by the feature extractor can be correctly identified by the classifier. Achieving this requires ensuring that the model can focus on non-erased areas with highly discriminative information (such as the part marked by the blue box). Through continuous adversarial gaming between the two, we can mine as much discriminative information as possible, thus protecting the re-ID model from the disruption caused by missing information due to occlusion. Similarly, our transforming defense and noising defense modules can also resist the interference of position misalignment and noisy information.

\subsection{Training and Inference}
During the training stage, we utilize the baseline learning module to extract discriminative features and three defense modules to enhance the occlusion robustness of the features. Correspondingly, the total loss can be formalized as:
\begin{equation}
\mathcal{L}_{total}=\mathcal{L}_{ce}+\lambda_{1}\mathcal{L}_{adv}^{e}+\lambda_{2}\mathcal{L}_{adv}^{t}+\lambda_{3}\mathcal{L}_{adv}^{n},
\end{equation}
where $\mathcal{L}_{adv}^{e}={\operatorname{argmax}}\mathcal{L}_{ce}^{e}(\boldsymbol W)+{\operatorname{argmin}}\mathcal{L}_{ce}^{e}(\boldsymbol E)$, and $\mathcal{L}_{adv}^{t}$ and $\mathcal{L}_{adv}^{n}$ are similar. $\lambda_{1}$, $\lambda_{2}$, and $\lambda_{3}$ are hyper-parameters utilized to control the relative importance of the three adversarial losses. The training process is carried out in an end-to-end manner, which is summarized in Algorithm \ref{Algo:1}.

In the test phase, we extract the feature map $F$ and process it with pooling and normalization. After that, the obtained feature vector is used for inference.

\section{Experiments}
\label{section5}
In this section, we conduct extensive experiments to demonstrate the effectiveness and superiority of the proposed method. First, we introduce six public person re-ID datasets used in the experiments. And then, we elaborate on the implementation details and evaluation protocols. After that, we compare the recognition performance of our approach with the state-of-the-art methods. Next, we verify the effectiveness of each module and analyze the impact of the hyper-parameters on the recognition rate. Finally, we further evaluate the proposed method with a discussion part.

\begin{algorithm}[t!]
    \caption{Training process of the proposed ETNDNet}
    \begin{algorithmic}[1]
        \REQUIRE A person image $x$ and its corresponding ground-truth $y$.
        \ENSURE Trained re-ID model. 
        \FOR {$ i = 1 $; $ i < iteration $; $ i ++ $ }
            \STATE {Extract $F\in \mathbb{R}^{H\times W\times C}$ by feature extractor $\boldsymbol E$.}
            \STATE {Initialize regions $R_{e}$, $R_{t_{1}}$, $R_{t_{2}}$, and $R_{n}$.}
             \STATE {Set each pixel in $R_{e}$ of $F$ to 0 to generate adversarial representation with incomplete information $F_{e}$.}
             \STATE {Transform pixels in $R_{t_{1}}$ to $R_{t_{2}}$ of $F$ to generate adversarial representation with misaligned information $F_{t}$.}
            \STATE {Replace each pixel in $R_{n}$ of $F$ with random value to generate adversarial representation with noisy information $F_{n}$.}
            \STATE {Send $F$, $F_{e}$, $F_{t}$ and $F_{n}$ to identity classifier $\boldsymbol W$ for identifying.}
            \STATE {Optimize $\boldsymbol W$ by Eq.(\textcolor{red}{1}), Eq.(\textcolor{red}{5}), Eq.(\textcolor{red}{7}), and Eq.(\textcolor{red}{9}).}
            \STATE {Optimize $\boldsymbol E$ by Eq.(\textcolor{red}{1}), Eq.(\textcolor{red}{6}), Eq.(\textcolor{red}{8}), and Eq.(\textcolor{red}{10}).}
            
        \ENDFOR
    \end{algorithmic} 
    \label{Algo:1}
\end{algorithm}

\subsection{Datastes}
\textbf{Occluded-DukeMTMC} \cite{PGFA} is a classical occluded person database derived from DukeMTMC-reID \cite{Duke}. It contains 15,618 training images of 702 pedestrians and 19,871 testing images of 1,110 pedestrians. Due to various occlusions in both query and gallery images, this dataset is quite challenging.

\textbf{Occluded-REID} \cite{OccludedReID} contains 2,000 images of 200 identities, and each identity has five occluded images and five full-body images. Following \cite{OccludedReID, IGOAS}, we take occluded images as the queries and non-occluded images as the galleries, and randomly select half of them for training and the rest for testing.

\textbf{P-DukeMTMC-reID} \cite{OccludedReID} is another variant of DukeMTMC-reID that targets occluded person re-ID task. It consists of 12,927 training images of 665 identities and 11,216 testing images of 634 identities (2,163 occluded images in the query set and 9,053 holistic images in the gallery set).

\textbf{Market-1501} \cite{Market} consists of 12,936 training images, 3,368 query images, and 19,732 gallery images. All images are captured from 6 non-overlapping cameras deployed at the campus. This dataset primarily targets the holistic person re-ID task due to few images being occluded.

DukeMTMC-reID is widely used to evaluate the performance of holistic person re-ID, which contains 36,411 images of 1,812 pedestrians collected from 8 cameras (408 distractor identities only appear in one camera view). All images are divided into 16,522 training samples and 19,889 test samples (This dataset is for research use only in this work).

\textbf{MSMT17} \cite{PTGAN} includes 126,441 images of 4,101 pedestrians, in which 32,621 images of 1,041 identities are used for training and the remaining are used for testing. This dataset is currently the most challenging holistic person database due to the long time span of sample collection.

More details about each dataset can be found in Table \ref{Tab:1}.

\begin{table}[!t]\small
\centering {\caption{Details of different person re-ID datasets. 'ID': Number of identities. 'IMGS': Number of images. 'O-Duke': Occluded-DukeMTMC. 'P-Duke': P-DukeMTMC-reID}\label{Tab:1}
\renewcommand\arraystretch{1.2}
\begin{tabular}{c|c|cc}
\hline
 \hline
  \multirow{2}*{Datasets} & Train & \multicolumn{2}{c}{Test(ID/Imgs)} \\
 & (ID/Imgs) & Query & Gallery\\
  \hline
  
  O-Duke & 702/15,618 & 519/2,210 & 1,110/17,661 \\

  Occluded-REID  & 100/1,000 & 100/500 & 100/500 \\

  P-Duke & 665/12,927 & 634/2,163 & 634/9,053 \\
  
  Market-1501  & 751/12,936 & 750/3,368 & 750/19,732 \\

  DukeMTMC-reID & 702/16,522 & 702/2,228 & 1,110/17,661 \\

  MSMT17 & 1,041/32,621 & 3,060/11,659 & 3,060/82,161 \\
  
  \hline\hline
\end{tabular}}\vspace{-1mm}
\end{table}

\begin{table*}[!ht]\small
\centering {\caption{Performance comparison with state-of-the-art methods on Occluded-DukeMTMC and Occluded-REID. The best results of the two kinds of methods are marked in blue, and ours are marked in bold. 'Ext' represents external tools, 'Bac' indicates visual backbone ('CNN' denotes Convolutional Neural Networks and 'VIT' represents Vision Transformer). $^{\S}$ denotes the modified version of models that do not use external tools. ‘*’ indicates that Non-Local is incorporated into our backbone. '()' indicates the performance when treating Market-1501 as the training set. '-' denotes that no reported result is available.}\label{Tab:2}
\renewcommand\arraystretch{1.2}
\begin{tabular}{c|c|c|c|cccc|cccc}
\hline
 \hline
  \multirow{2}*{Methods} & \multirow{2}*{Ref} & \multirow{2}*{Ext} & \multirow{2}*{Bac} & \multicolumn{4}{c|}{Occluded-DukeMTMC} & \multicolumn{4}{c}{Occluded-REID}\\
\cline{5-12} & & & & Rank-1 & Rank-5 & Rank-10 & mAP & Rank-1 & Rank-5 & Rank-10 & mAP\\
  \hline

  PGFA \cite{PGFA} & ICCV'19 & \Checkmark & CNN & 51.4 & 68.6 & 74.9 & 37.3 & 57.1 & 77.9 & 84.0 & 56.2 \\
  
  PVPM \cite{PVPM} & CVPR'20 & \Checkmark & CNN & 47.0 & - & - & - & 70.4 & \textcolor{blue}{84.1} & \textcolor{blue}{89.8} & 64.2 \\

  HOReID \cite{HOReID} & CVPR'20 & \Checkmark & CNN & 55.1 & - & - & 43.8 & 80.3 & - & - & 70.2 \\

  GASM \cite{GASM} & ECCV'20 & \Checkmark & CNN & - & - & - & - & 74.5 & - & - & 65.6 \\

  OAMN \cite{OAMN} & ICCV'21 & \Checkmark & CNN & 62.6 & 77.5 & - & 46.1 & - & - & - & - \\

  Pirt \cite{PIRT} & ACMM'21 & \Checkmark & CNN & 60.0 & - & - & 50.9 & - & - & - & - \\

  PGFL-KD \cite{PGFL-KD} & ACMM'21 & \Checkmark & CNN & 63.0 & - & - & 54.1 & \textcolor{blue}{80.7} & - & - & \textcolor{blue}{70.3} \\

  PGMANet \cite{PGMANet} & IJCNN'21 & \Checkmark & CNN & 51.3 & 66.5 & 73.4 & 40.9 & - & - & - & - \\

  RFCnet \cite{RFC} & TPAMI'21 & \Checkmark & CNN & \textcolor{blue}{63.9}& \textcolor{blue}{77.6} & \textcolor{blue}{82.1}& \textcolor{blue}{54.5} & - & - & - & - \\

  PEFB \cite{PEFB} & TNNLS'21 & \Checkmark & CNN & 56.3 & 72.4 & 78.0 & 43.5 & - & - & - & - \\

  SORN \cite{SORN} & TCSVT'21 & \Checkmark & CNN & 57.6 & 73.7 & 79.0 & 46.3 & - & - & - & - \\

    \hline

  Adver Occluded \cite{AdverOccluded} & CVPR'18 & \XSolidBrush & CNN & 44.5 & - & - & 32.2 & - & - & - & - \\

  PCB \cite{PCB} & ECCV'18 & \XSolidBrush & CNN & 42.6 & 57.1 & 62.9 & 33.7 & 66.6 & 89.2 & - & - \\

  AFPB \cite{OccludedReID} & ICME'18 & \XSolidBrush & CNN & - & - & - & - & 68.1 & 88.3 & 93.7 & - \\

  FPR \cite{FPR} & ICCV'19 & \XSolidBrush & CNN & - & - & - & - & 78.3 & - & - & 68.0 \\

  ISP \cite{ISP} & ECCV'20 & \XSolidBrush & CNN & 62.8 & 78.1 & 82.9 & 52.3 & - & - & - & - \\

  RE \cite{RE} & AAAI'20 & \XSolidBrush & CNN & 40.5 & 59.6 & 66.8 & 30.0 & 65.8 & 87.9 & - & - \\

  RFCnet$^{\S}$ \cite{RFC} & TPAMI'21 & \XSolidBrush & CNN & 52.4 & 68.3 & 73.4 & 44.8 & - & - & - & - \\

  IGOAS \cite{IGOAS} & TIP'21 & \XSolidBrush & CNN & 60.1 & - & - & 49.4 & 81.1 & 91.6 & - & - \\

  ASAN \cite{ASAN} & TCSVT'21 & \XSolidBrush & CNN & 55.4 & 72.4 & 78.9 & 43.8 & 82.5 & 92.2 & - & 71.8 \\

  CBDBNet \cite{CBDBNet} & TCSVT'21 & \XSolidBrush & CNN & 50.9 & 66.0 & 74.2 & 38.9 & - & - & - & - \\

  FED \cite{FED} & CVPR'22 & \XSolidBrush & VIT & \textcolor{blue}{68.1} & 79.3 & - & \textcolor{blue}{56.4} & \textcolor{blue}{86.3} & - & - & \textcolor{blue}{79.3} \\


  QPM \cite{QPM} & TMM'22 & \XSolidBrush & CNN & 64.4 & 79.3 & 84.2 & 49.7 & - & - & - & - \\

  DRL-Net \cite{DRLNet} & TMM'22 & \XSolidBrush & VIT & 65.8 & \textcolor{blue}{80.4} & \textcolor{blue}{85.2} & 53.9 & - & - & - & - \\

  OPR-DAAO \cite{OPR-DAAO} & TIFS'22 & \XSolidBrush & CNN & 66.2 & 78.4 & 83.9 & 55.4 & 84.2 & 87.3 & - & 75.1\\

    RTGTA \cite{RTGAT} & TIP'23 & \XSolidBrush & CNN & 61.0 & 69.7& 73.6 & 50.1 & 71.8 & \textcolor{blue}{94.6} & \textcolor{blue}{99.0} & 51.0\\

    \hline

    \bf{Baseline} & - & \XSolidBrush & CNN & \bf{52.5} & \bf{69.3} & \bf{75.0} & \bf{44.6} & \bf{87.8} & \bf{95.3} & \bf{97.5} & \bf{79.5} \\

    \bf{Ours} & - & \XSolidBrush & CNN & \bf{63.6} & \bf{79.5} & \bf{84.7} & \bf{54.7} & \bf{89.3} & \bf{96.0} & \bf{97.9} & \bf{81.5} \\

    \bf{Ours*} & - & \XSolidBrush & CNN & \bf{68.1} & \bf{82.5} & \bf{87.7} & \bf{57.6} & \bf{90.5} & \bf{96.5} & \bf{98.5} & \bf{81.9} \\
  
  \hline\hline
\end{tabular}}
\vspace{-1mm}
\end{table*}

\subsection{Settings}
\textbf{Implementation Details}: We conduct all experiments on one GTX3090 with the Pytorch platform. All images are uniformly resized to 256$\times$128 and subjected to random flipping, padding, cropping, and erasing to achieve data augmentation. The batch size is set to 64, each batch contains 8 pedestrians and each pedestrian has 8 images. We set the initial learning rate to $3\times10^{-4}$ and decrease it by a factor of 0.1 at the 40th epoch and 70th epoch. The size of the feature map is $H=16$, $W=8$, and $C=2048$. The hyper-parameters are set to $\lambda_{1}=0.1$, $\lambda_{2}=0.15$, and $\lambda_{3}=0.1$. We utilize the Adam optimizer \cite{Adam} to update the model parameters and train 120 epochs in total.

\begin{table}[!t]\small
\centering {\caption{Performance comparison with the state-of-the-art methods on P-DukeMTMC-reID. }\label{Tab:3}
\renewcommand\arraystretch{1.2}
\begin{tabular}{c|c|cccc}
\hline
 \hline
  Methods & Bac & Rank-1 & Rank-5 & Rank-10 & mAP \\
  \hline
  
  PCB \cite{PCB} & CNN & 79.4 & 87.1 & 90.0 & 63.9 \\

  IDE \cite{IDE} & CNN & 82.9 & 89.4 & 91.5 & 65.9 \\

  PVPM \cite{PVPM} & CNN & 85.1 & 91.3 & 93.3 & 69.9 \\

  PGFA \cite{PGFA} & CNN & 85.7 & 92.0 & 94.2 & 72.4 \\

  ISP \cite{ISP} & CNN & \textcolor{blue}{89.0} & \textcolor{blue}{94.1} & \textcolor{blue}{95.3} & \textcolor{blue}{74.7} \\
  
  \hline
  \textbf{Baseline} & CNN & \bf{91.2} & \bf{95.0} & \bf{95.9} & \bf{76.5} \\
  \textbf{Ours} & CNN & \bf{91.6} & \bf{95.1} & \bf{96.2} & \bf{79.2} \\
  \textbf{Ours*} & CNN & \bf{92.7} & \bf{95.6} & \bf{96.6} & \bf{80.7} \\
  \hline\hline
\end{tabular}}%
\vspace{-1mm}
\end{table}

\textbf{Evaluation Protocol}: We follow the public evaluation indicators to verify the performance of the proposed method, namely Cumulative Matching Characteristics (CMC) and mean Average Precision (mAP). Note that all experiments are performed in the single query setting.

\subsection{Comparison with State-of-the-art Methods}

\textbf{Experiments on occluded datasets}: We first evaluate the superiority of the proposed method on the three public occluded datasets, namely \textbf{Occluded-DukeMTMC}, \textbf{Occluded-REID}, and \textbf{P-DukeMTMC-reID}. The comparison results are shown in Table \ref{Tab:2} and Table \ref{Tab:3}, where we report the accuracy of Rank-1, Rank-5, Rank-10, and mAP.

For Occluded-DukeMTMC, methods that rely on external models have achieved satisfactory results. For example, RFCNet \cite{RFC} achieves 63.9\% Rank-1 and 54.5\% mAP by utilizing pose and semantic information. However, this superior accuracy is dependent on the reliability of external models and requires significant computational resources. As shown in the table, in the absence of auxiliary information, RFCNet only achieves 52.4\% (-11.5\%) Rank-1 and 44.8\% (-9.7\%) mAP. These results demonstrate the heavy reliance of such methods on external visual cues. In contrast, our method achieves comparable performance to RFCNet without utilizing external models and with minimal computational resources. Recently, model-free methods without external models have gained attention, and our proposed method falls into this category while having notable advantages in terms of performance and model simplicity. Specifically, the optimal ISP \cite{ISP} predicts human body parts using clustering, avoiding the need for a pre-trained semantic parsing network, but it suffers from significant computational complexity and inferior performance compared to our method (52.3\% vs. 54.7\% mAP). Furthermore, Adver Occluded \cite{AdverOccluded} and IGOAS \cite{IGOAS}, which utilize adversarial defense by generating aggressive samples, achieve only 44.5\% and 60.1\% Rank-1 accuracy, and 32.2\% and 49.4\% mAP, respectively, while our ETNDNet achieves 63.6\% Rank-1 accuracy and 54.7\% mAP. These results highlight the stronger occlusion defense ability of the model trained by our method. Additionally, the attention mechanism is commonly employed by state-of-the-art methods \cite{FED, DRLNet} to extract feature maps containing crucial pedestrian information. In our proposed ETNDNet, adversarial representations formed by perturbing such feature maps are more aggressive, and leveraging them for adversarial defense enables the re-ID model to learn more robust features against occlusions. To achieve this, we incorporate Non-Local \cite{Non-Local} into our backbone. As shown in Table \ref{Tab:2}, our method (Ours*) achieves superior performance, surpassing the aforementioned methods (e.g., +1.2\% mAP compared to FED \cite{FED}), further demonstrating the effectiveness and superiority of our approach.

For Occluded-REID, we repeat the experiment 10 times as in \cite{IGOAS} and take the mean as the final result. As shown in Table \ref{Tab:2}, the proposed method achieves satisfactory results. For example, compared with the optimal pose-guided based method PGFL-KD \cite{PGFL-KD}, the proposed ETNDNet improves the Rank-1 accuracy from 80.7\% to 89.3\%, and mAP from 70.3\% to 81.5\%. Compared with the optimal external model-free method FED \cite{FED}, Rank-1 and mAP of ours are improved by 3.0\% and 2.2\%. Moreover, IGOAS \cite{IGOAS}, which is based on adversarial learning, the proposed ETNDNet surpasses it by 8.2\% in Rank-1 accuracy. Finally, for P-DukeMTMC-reID,  as shown in Table \ref{Tab:3}, our method also outperforms state-of-the-art algorithms and achieves significant improvements in mAP, such as 6.8\% over the optimal ISP \cite{ISP}.

According to the above results, the proposed method has significant advantages for occluded person re-ID, which is mainly attributed to the fact that our adversarial representations fully simulate perturbation caused by various occlusion issues (missing information, position misalignment, and noisy information), and the model trained by the developed GAN-based adversarial defense has the ability to protect the re-ID system from such disturbance, effectively overcoming the adverse effect of occlusion issues on recognition performance.

\textbf{Experiments on holistic datasets}: To further evaluate the superiority and practicality of our method, we compare with the state-of-the-art methods on holistic re-ID datasets, i.e., \textbf{Market-1501}, \textbf{DukeMTMC-reID}, and \textbf{MSMT17}, and report the comparison results in Table \ref{Tab:4}.

\begin{table*}[!ht]\small
\centering {\caption{Performance comparison with state-of-the-art methods on Market-1501, DukeMTMC-reID, and MSMT17.}\label{Tab:4}
\renewcommand\arraystretch{1.2}
\begin{tabular}{c|c|ccc|ccc|ccc}
\hline
 \hline
  \multirow{2}*{Methods} & \multirow{2}*{Bac} & \multicolumn{3}{c|}{Market-1501} & \multicolumn{3}{c|}{DukeMTMC-reID} & \multicolumn{3}{c}{MSMT17}\\
\cline{3-11} & & Rank-1 & Rank-5 & mAP & Rank-1 & Rank-5 & mAP & Rank-1 & Rank-5 & mAP \\
  \hline

  PTGAN \cite{PTGAN} & CNN & - & - & - & - & - & - & 68.2 & - & 40.4 \\

  PGFA \cite{PGFA} & CNN & 91.2 & - & 76.8 & 82.6 & - & 65.5 & - & - & - \\

  PEFB \cite{PEFB} & CNN & 92.7 & - & 81.3 & 86.2 & - & 72.6 & - & - & - \\
  
  VPM \cite{VPM} & CNN & 93.0 & 97.8 & 80.8 & 83.6 & 91.7 & 72.6 & - & - & - \\

  IGOAS \cite{IGOAS} & CNN & 93.4 & - & 84.1 & 86.9 & - & 75.1 & - & - & - \\
  
  HOReID \cite{HOReID} & CNN & 94.2 & - & 84.9 & 86.9 & - & 75.6 & - & - & - \\
  
  IANet \cite{IANet} & CNN & 94.4 & - & 83.1 & 87.1 & - & 73.4 & 75.5 & \textcolor{blue}{85.5} & 46.8 \\
  
  BagTricks \cite{BAGTRICKS} & CNN & 94.5 & - & 85.9 & 86.4 & -& 76.4 & - & - & - \\
  
  MHSANet \cite{MHSANet} & CNN & 94.6 & - & 84.0 & 87.3 & - & 73.1 & - & - & - \\
  
  SORN \cite{SORN} & CNN & 94.8 & - & 84.5 & 86.9 & - & 74.1 & - & - & - \\

  OSNet \cite{OSNet} & CNN & 94.8 & - & 84.9 & 88.6 & - & 73.5 & \textcolor{blue}{78.7} & - & 52.9 \\

  AGW \cite{AGW} & CNN & 95.1 & - & 87.8 & 89.0 & - & \textcolor{blue}{79.6} & 78.3 & - & \textcolor{blue}{55.6} \\

  POS \cite{POS} & CNN & 95.0 & 98.3 & 86.2 & 94.6 & 88.7 & 76.7 & - & - & - \\

  PLIP \cite{PLIP} & CNN & 95.1 & - & \textcolor{blue}{88.0} & 86.5 & - & 77.0 & - & - & - \\

  BPBreID \cite{BPBreID} & CNN & 95.1 & - & 87.0 & \textcolor{blue}{89.6} & - & 78.3 & - & - & - \\

  GPS \cite{GPS} & CNN & \textcolor{blue}{95.2} & \textcolor{blue}{98.4} & 87.8 & 88.2 & \textcolor{blue}{95.2} & 78.7 & - & - & - \\

  \hline
  \textbf{Baseline} & CNN & \bf{93.8} & \bf{98.1} & \bf{84.1} & \bf{86.1} & \bf{93.2} & \bf{74.2} & \bf{74.8} & \bf{85.2} & \bf{49.0} \\
  \textbf{Ours} & CNN & \bf{95.3} & \bf{98.2} & \bf{87.2} & \bf{88.5} & \bf{94.7} & \bf{77.9} & \bf{80.9} & \bf{89.3} & \bf{56.6}\\
  \textbf{Ours*} & CNN & \bf{95.7} & \bf{98.5} & \bf{88.7} & \bf{89.3} & \bf{95.3} & \bf{78.8} & \bf{82.7} & \bf{91.1} & \bf{58.0} \\
  \hline\hline
\end{tabular}}
\vspace{-0.3cm}
\end{table*}

Although these three datasets are released for holistic person re-ID, due to the changeable postures and complex backgrounds, they still exist problems of information loss, information misalignment, and noisy information. Therefore, our method is also applicable and can further improve the recognition rate. Specifically, for Market-1501, our ETNDNet outperforms the state-of-the-art GPS \cite{GPS} in terms of Rank-1 accuracy. Additionally, the integration of an attention mechanism into our framework led to a further improvement in recognition performance, with Rank-1 being 0.5\% higher than GPS and mAP being 0.7\% higher than PLIP \cite{PLIP}. As for DukeMTMC-reID, our ETNDNet also surpasses most state-of-the-art methods. Moreover, MSMT17 is currently the largest dataset, and it is relatively difficult due to the large difference between positive and negative pairs. Existing methods increase training samples (PTGAN \cite{PTGAN}) or introduce attention mechanisms (IANet \cite{IANet}, OSNet \cite{OSNet}, and AGW \cite{AGW}) to reduce the intra-class distance. Even so, our method with less training cost achieves superior performance with 80.9\% Rank-1 and 56.6\% mAP,  which is ahead of the state-of-the-art methods.

\subsection{Model Analysis}

\textbf{Ablation Studies}. We conduct ablation experiments to demonstrate the contribution of each module in the proposed ETNDNet, including erasing defense (ED), transforming defense (TD), and noising defense (ND). All experiments were performed on Occluded-DukeMTMC, which can better reflect the effectiveness of modules. The Rank-1 accuracy and mAP (\%) are reported in Table \ref{Tab:5}.

\textit{Effectiveness of ED.} The developed ED is designed to protect the re-ID system from missing information. As shown in Table \ref{Tab:5}, when deploying it to the baseline re-ID system, the Rank-1 accuracy and mAP are improved from 52.5\% to 58.7\% and from 44.6\% to 49.7\%, respectively. Moreover, compared with NDNet, the Rank-1 and mAP obtained by ENDNet have also been significantly improved. These results verify that our ED is indeed able to promote the re-ID model to learn features that are not disturbed by information loss, playing a positive role in identifying occluded target pedestrians and can cooperate with other modules to further improve the recognition performance.

\begin{figure*}[t!]
  \centering
  
  \subfigbottomskip=-1pt
  \subfigcapskip=-1pt
  \subfigure[$\lambda_{1}$] {\includegraphics[height=2.0in,width=2.4in,angle=0]{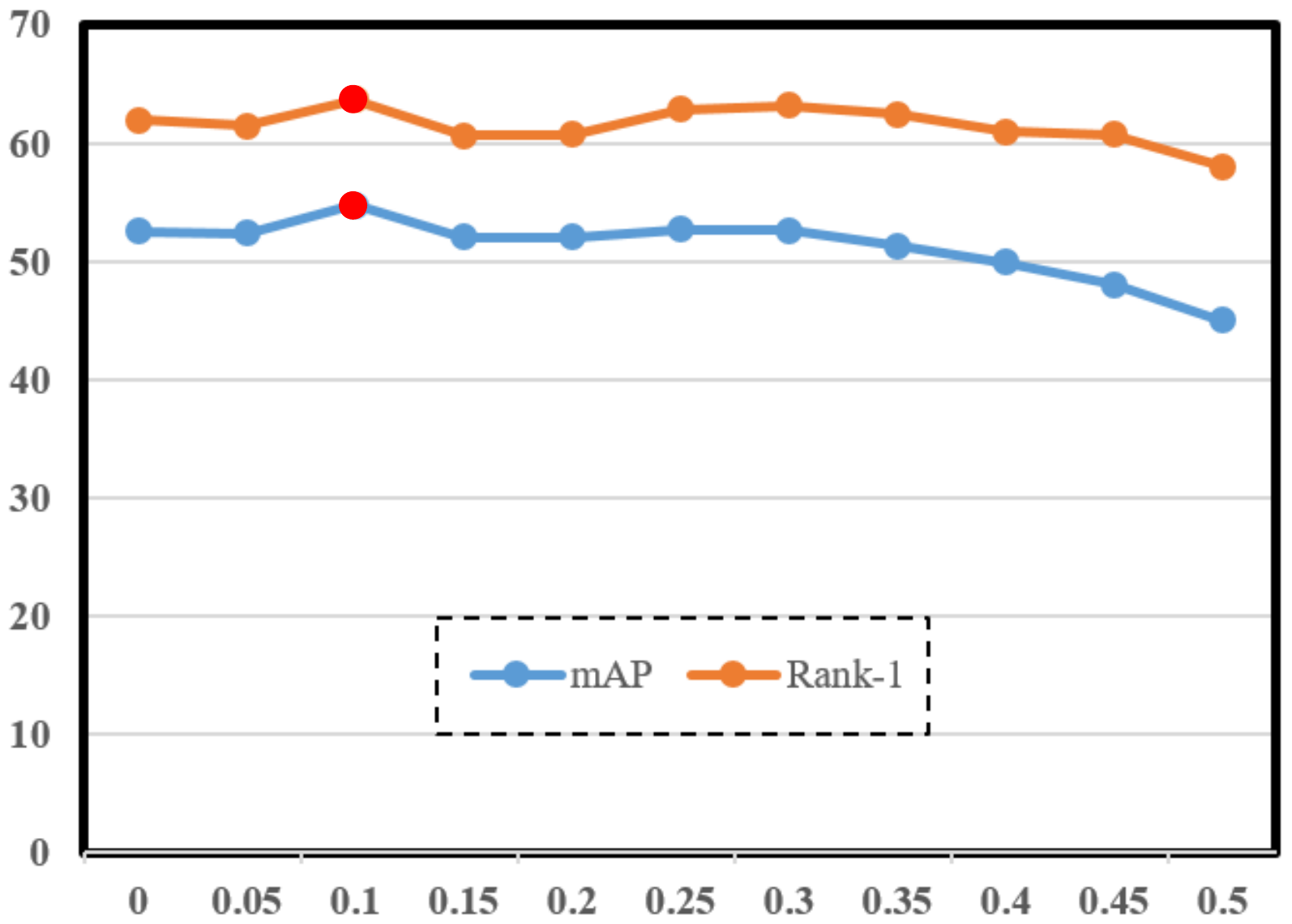}}
  \subfigure[$\lambda_{2}$] {\includegraphics[height=2.0in,width=2.4in,angle=0]{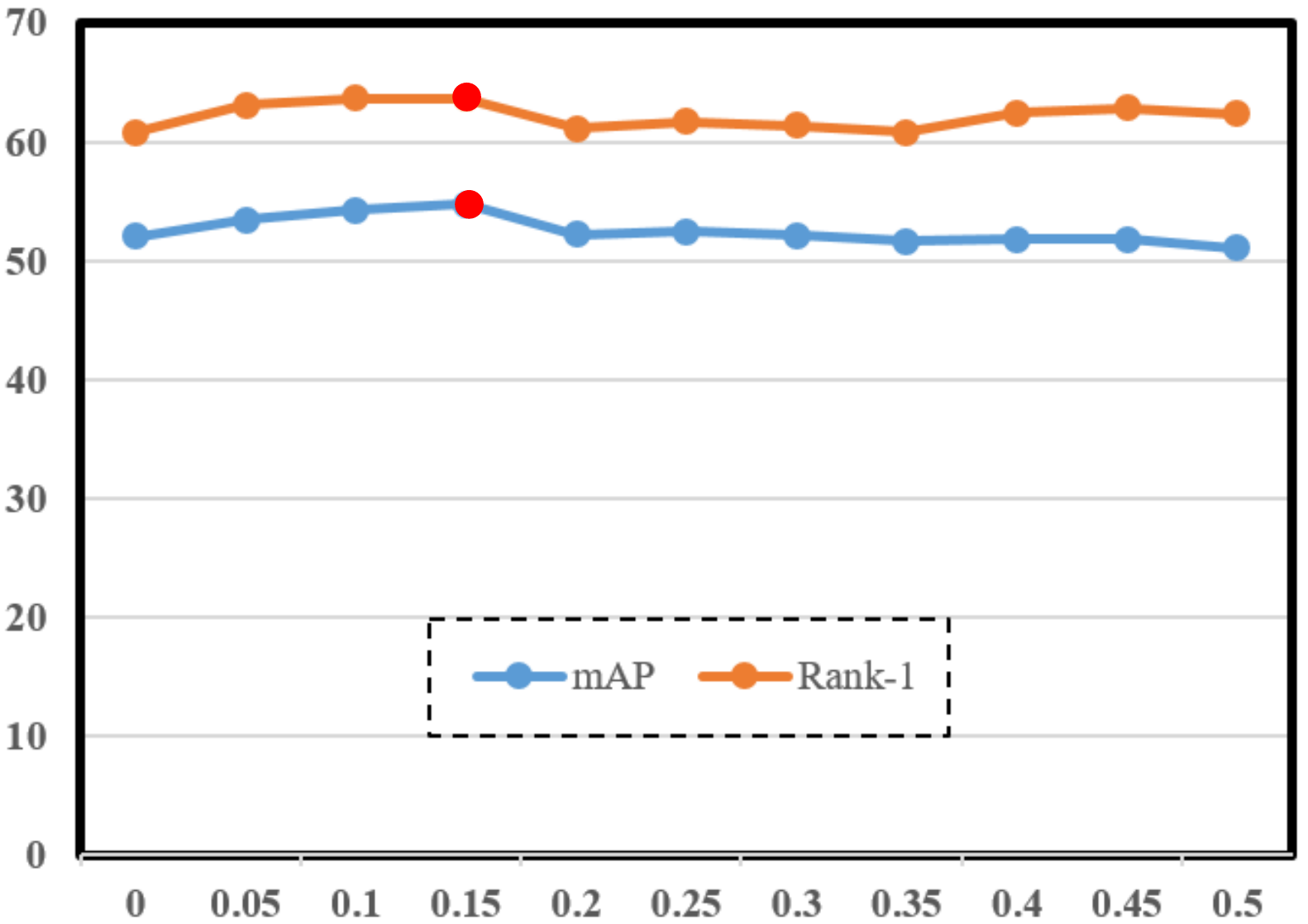}}
  \subfigure[$\lambda_{3}$] {\includegraphics[height=2.0in,width=2.4in,angle=0]{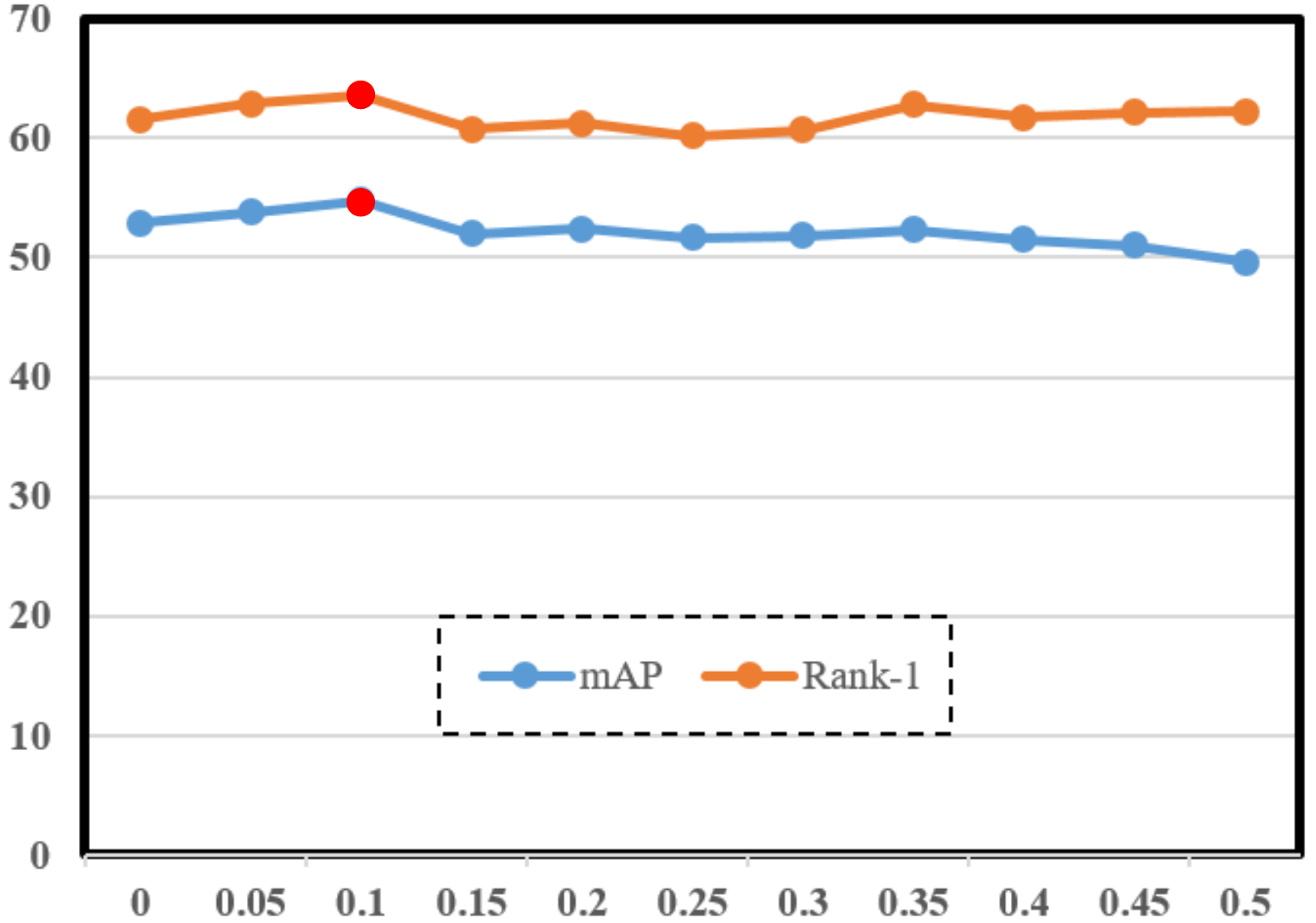}}\\
  \caption{The effect analysis on different hyper-parameters $\lambda_{1}$, $\lambda_{2}$, and $\lambda_{3}$. Rank-1 accuracy and mAP are reported and red dots represent the optimal values.}
  \label{Fig:4}
  \vspace{-0.3cm}
\end{figure*}

\textit{Effectiveness of TD.} The proposed TD strives to defend the re-ID model for extracting features unaffected by misaligned information. As shown in Table \ref{Tab:5}, our TDNet promotes the Rank-1 accuracy and mAP of the Baseline by a large margin. And, when it is equipped on the above EDNet, the Rank-1 accuracy and mAP are improved from 58.7\% to 62.1\% and from 49.7\% to 52.8\%. These results demonstrate that our TD is also beneficial to occluded person re-ID, effectively alleviating the adverse impact of position misalignment caused by occlusion on recognition performance.

\begin{table}[!ht]\small
\centering {\caption{Ablation studies of the proposed ETNDNet.}\label{Tab:5}

\renewcommand\arraystretch{1.2}
\begin{tabular}{c|cccc|cc}
\hline
 \hline
  Methods & B & ED & TD & ND & Rank-1 & mAP \\
  \hline
  
  Baseline & \checkmark &  &  &  & 52.5 & 44.6 \\

  EDNet & \checkmark & \checkmark &  &  & 58.7 & 49.7 \\

  TDNet & \checkmark &  & \checkmark &  & 57.5 & 49.7 \\

  NDNet & \checkmark &  &  & \checkmark & 58.0 & 48.8 \\

  ETDNet & \checkmark & \checkmark & \checkmark &  & 62.1  & 52.8 \\

  ENDNet & \checkmark & \checkmark &  & \checkmark & 62.3 & 52.6 \\

  TNDNet & \checkmark &  & \checkmark & \checkmark & 59.0 & 51.2 \\

  ETNDNet & \checkmark & \checkmark & \checkmark & \checkmark & \bf{63.6} & \bf{54.7} \\

  ETNDNet* & \checkmark & \checkmark & \checkmark & \checkmark & \bf{68.1} & \bf{57.6} \\

  \hline\hline
\end{tabular}}
\vspace{-0.2cm}
\end{table}

\textit{Effectiveness of ND.} The proposed ND is dedicated to solving the problem of occlusion noise introduced by obstacles and non-target pedestrians, which is ignored in existing studies. As shown in Table \ref{Tab:5}, it also contributes to the improvement of recognition performance, and when it is combined with other modules, the formed framework can achieve higher accuracy. Specifically, our NDNet increases the Rank-1 accuracy of the re-ID baseline model by 5.5\% and mAP by 4.2\%. And, the performance of ENDNet and TNDNet is also higher than that of EDNet and TDNet. These results confirm our ND is indeed robust against the interference of noisy information.

Finally, when all modules are equipped, the developed ETNDNet can achieve the best recognition rates of 63.6\% Rank-1 and 54.7\% mAP. It follows that our three adversarial defense modules are able to collaborate with each other, effectively overcoming the negative effects of perturbation of various occlusion issues on identification.

\textbf{Parameter Analysis}. Hyper-parameters $\lambda_{1}$, $\lambda_{2}$, and $\lambda_{3}$ are used to balance the relative importance of adversarial losses in the above three adversarial defense modules. Here, we study the influence of their different values on recognition performance. Each parameter was tested with 11 different values and the experimental results are illustrated in Figure \ref{Fig:4}. Note that when one of the parameters is analyzed, the remaining two are fixed at the optimal values.

\textit{The effect of $\lambda_{1}$.} From Figure \ref{Fig:4}, we can see that when $\lambda_{1}$ increases from 0 to 0.1, both Rank-1 accuracy and mAP are improved, and when $\lambda_{1}=0.1$, our method achieves the best recognition performance. This indicates that $\lambda_{1}=0.1$ is an optimal value for the proposed method. Moreover, we observe that the recognition performance decreases as $\lambda_{1}$ continues to increase, which may be due to over-guaranteeing the robustness of the learned features while destroying their discriminativeness.

\textit{The effect of $\lambda_{2}$.} The hyper-parameter $\lambda_{2}$ is used to control the relative importance of the adversarial loss in the proposed transforming defense module. As shown in Figure \ref{Fig:4}, the performance of the developed ETNDNet remains relatively stable when $\lambda_{2}$ increases from 0 to 0.5. And, when $\lambda_{2}=0.15$, the recognition performance slightly exceeds the corresponding performance achieved at other values of $\lambda_{2}$. So, $\lambda_{2}$ is set to 0.15 in all the experiments.

\textit{The effect of $\lambda_{3}$.} In order to further improve the robustness of the learned features, the ND is designed and the parameter {$\lambda_{3}$ is used to control its importance.  It can be seen that Rank-1 accuracy and mAP reach the peak when $\lambda_{3}=0.1$, which further confirms the validity of ND and the rationality of $\lambda_{3}=0.1$.

\subsection{Further Discussions}
In this section, we further evaluate the proposed method. All experiments were performed on Occluded-Duke with the same hardware (1 GTX3090 GPU, 8 CPU, and 32GB memory).

\textbf{Erasing Strategy}. In our erasing module, we perturb feature maps with a random erasing strategy to form adversarial representations with incomplete information. The strategy proposed in \cite{DropBlock} can also serve this purpose. However, since it fixes the erased area's size proportion and aspect ratio while different obstacles have various sizes and shapes in real scenes, perturbing feature maps by such strategy hinders the aggressiveness of adversarial representations and ultimately limits recognition performance improvement. As shown in Figure \ref{Fig:5}(a), when we follow DropBlock \cite{DropBlock} to fix the erased area's size proportion and aspect ratio to 0.3, the Rank-1 and mAP of our EDNet dropped by 1.3\% and 3.0\%. Obviously, it is more appropriate to adopt our random erasing strategy to perturb feature maps.

\begin{figure}[t!]
\centering
\subfigbottomskip=-1pt
\subfigcapskip=-1pt
\subfigure[] {\includegraphics[height=1.1in,width=1.7in,angle=0]{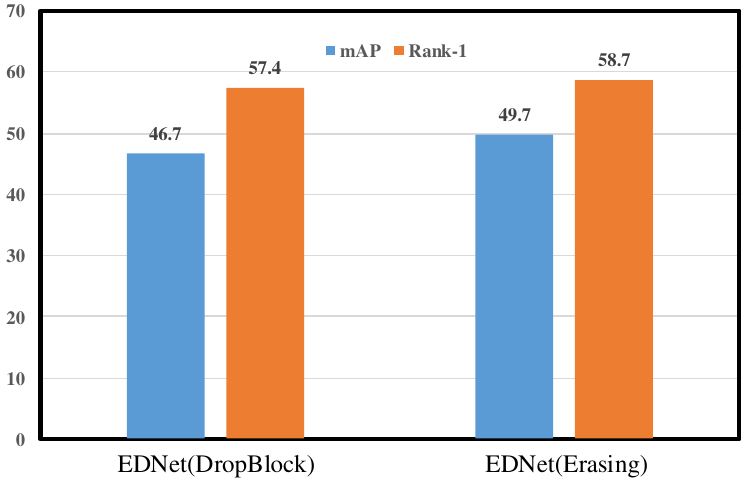}}
\subfigure[] {\includegraphics[height=1.1in,width=1.7in,angle=0]{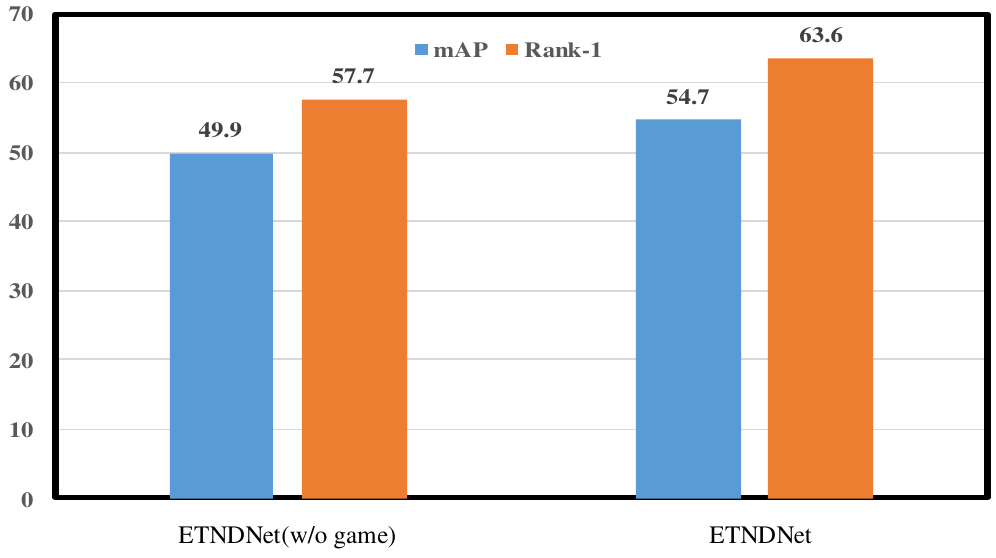}}
\caption{(a) The effect of DropBlock and our erasing strategy on the performance of our EDNet. (b) The performance impact of the adversarial game. ‘w/o game’ indicates the adversarial game is removed.}
\label{Fig:5}
\vspace{-0.3cm}
\end{figure}

\textbf{Defense Paradigm}. In this paper, we design a novel GAN-based adversarial defense paradigm to protect the re-ID system from various occlusion disturbances, which is significantly different from the adversarial representation learning of IGOAS \cite{IGOAS}. Specifically, IGOAS employs an adversarial suppression branch that implements a mean square loss between the feature map and its masked version to make the model focus on pedestrian information in non-erased areas. In contrast, our GAN-based adversarial defense paradigm continuously plays an adversarial game between the identity classifier and feature extractor, encouraging the model to focus on highly discriminative information in non-disturbed areas. Particularly, in the developed defense paradigm, optimizing the feature extractor alone without the adversarial game process is comparable to IGOAS's learning approach (that is, treating adversarial representations as supervision signals to optimize the feature extractor, making the model focus on pedestrian information in non-disturbed areas). As demonstrated in Figure \ref{Fig:5}(b), removing the adversarial game process reduces the mAP and Rank-1 accuracy by 4.8\% and 5.9\%, respectively, indicating the superiority of the learning way in our GAN-based adversarial defense paradigm.

\textbf{Model Complexity}. To further demonstrate the practicality of the proposed method, we conduct an analysis of the model complexity and compare ours with four state-of-the-art methods. The results are illustrated in Table \ref{Tab:6}, where we report the parameter number of models (Param), the number of floating-point operations for an input image (FLOPs), and the inference time of retrieving all queries (IT). It can be observed that existing state-of-the-art methods have a relatively large model complexity. In comparison, our Param, FLOPs, and IT are only 24.9M, 4.1, and 25s, thus the proposed ETNDNet has a significant advantage in model simplicity. Moreover, our method does not introduce any external modules with learnable parameters. Therefore, it has the same model complexity as the baseline, that is to say, our method improves the recognition performance without increasing the computational burden, with strong practicability. Additionally, incorporating Non-Local into our backbone, ETNDNet* increases the number of parameters by only 0.2M, but leads to significant performance improvements, surpassing state-of-the-art methods.

\begin{table}[!t]\small
\centering {\caption{Model complexity analysis. Param: the parameter number of models. FLOPs: the number of floating-point operations for an input image. ‘IT’: The inference time of retrieving all queries.}\label{Tab:6}
\renewcommand\arraystretch{1.2}
\begin{tabular}{c|c|c|c|c|c}
\hline
 \hline
  Methods & Param & FLOPs & IT & Rank-1 & mAP \\
  \hline

  HOReID \cite{HOReID} & 163.2M & 4.9 & 258s & 55.1 & 43.8 \\

  IGOAS \cite{IGOAS} & 32.6M & 4.6 & 49s & 60.1 & 49.4 \\

  RFCnet \cite{RFC} & 25.6M & 4.3 & 119s & 63.9 & 54.5 \\

  OPR-DAAO \cite{OPR-DAAO} & 58.2M & 9.2 & - & 66.2 & 55.4 \\

  FED \cite{FED} & 146.2M & 11.0 & 98s & 68.1 & 56.4 \\
  
  \hline
  \textbf{Ours(Baseline)} & \bf{24.9M} & \bf{4.1} & 25s & \bf{52.5} & \bf{44.6} \\
  \textbf{Ours(ETNDNet)} & \bf{24.9M} & \bf{4.1} & 25s & \bf{63.6} & \bf{54.7} \\
  \textbf{Ours(ETNDNet*)} & \bf{25.1M} & \bf{4.1} & 44s & \bf{68.1} & \bf{57.6} \\
  \hline\hline
\end{tabular}}
\end{table}

\textbf{Model Generalization Ability.} We also evaluate the model generalization ability of our algorithms, with the results presented in Table \ref{Tab:7}. Importantly, it should be noted that our ETNDNet was not primarily tailored for domain generalization, and its performance falls short of the state-of-the-art method \cite{SNR}. Nonetheless, in comparison to other occluded person re-ID algorithms, our model demonstrates superior generalization ability. To illustrate, for Market $\rightarrow$ Duke, our ETNDNet outperforms HOReID by 1.6\% and IGOAS by 0.2\% in terms of mAP. Furthermore, compared with the Baseline, our algorithm yields a substantial enhancement of 6.0\% in Rank-1 accuracy and 4.1\% in mAP. This observation underscores the efficacy of the proposed algorithm in enhancing model generalization.

\begin{table}[!ht]\small
\centering {\caption{Performance comparison of Model generalization ability. A $\rightarrow$ B denotes the model is trained on dataset A and directly tested on dataset B.}\label{Tab:7}
\renewcommand\arraystretch{1.2}
\begin{tabular}{c|cc|cc}
\hline
 \hline
  \multirow{2}*{Methods} & \multicolumn{2}{c|}{Market $\rightarrow$ Duke} & \multicolumn{2}{c}{Duke $\rightarrow$ Market} \\
\cline{2-5} & Rank-1 & mAP & Rank-1 & mAP\\
  \hline

  QAConv \cite{QAConv} & 48.8 & 28.7 & 58.6 & 27.2 \\

  SNR \cite{SNR} & 55.1 & 33.6 & 66.7 & 33.9 \\

  \hline

  HOReID \cite{HOReID} & 39.4 & 23.2 & 48.0 & 22.1 \\

  IGOAS \cite{IGOAS} & 41.5 & 24.6 & 48.7 & 22.2 \\
  
  \hline
  \textbf{Baseline} & \bf{35.1} & \bf{20.7} & \bf{56.2} & \bf{26.9} \\
  \textbf{Ours} & \bf{41.1} & \bf{24.8} & \bf{56.3} & \bf{27.6} \\
  \textbf{Ours*} & \bf{48.3} & \bf{30.6} & \bf{60.1} & \bf{30.2} \\
  \hline\hline
\end{tabular}}
\vspace{-0.1cm}
\end{table}

\textbf{Adversarial Representation}. Aggressiveness is an important basis for judging the quality of adversarial examples. As discussed earlier, our adversarial representations formed by perturbing feature maps are more aggressive than adversarial samples formed in \cite{AdverOccluded, IGOAS}. To demonstrate this, we test the performance of the model trained by the baseline framework on adversarial samples and various adversarial representations. The experimental results are shown in Figure \ref{Fig:6} (a). Firstly, compared with the baseline, the adversarial samples reduce the mAP and Rank-1 by 9.3\% and 8.0\%, while our adversarial representations with incomplete information reduce the mAP and Rank-1 by 20.8\% and 19.1\%, which suggests that applying perturbation on the feature map is indeed more aggressive. Secondly, the adversarial representations generated by random transforming simulate the issue of position misalignment caused by occlusion, which is a capability that adversarial samples do not possess. And, as can be seen from Figure \ref{Fig:6} (a), they lead to the reduction of the mAP and Rank-1 from 44.6\% to 36.0\% and from 52.5\% to 46.1\%. Thirdly, our adversarial representations with noisy information also reduce the recognition rate by a large margin (-16.3\% mAP and -16.5\% Rank-1) and are more effective in attacking the re-ID model than adversarial samples.

\begin{figure}[t!]
\centering
\subfigbottomskip=-1pt
\subfigcapskip=-1pt
\subfigure[] {\includegraphics[height=1.1in,width=1.7in,angle=0]{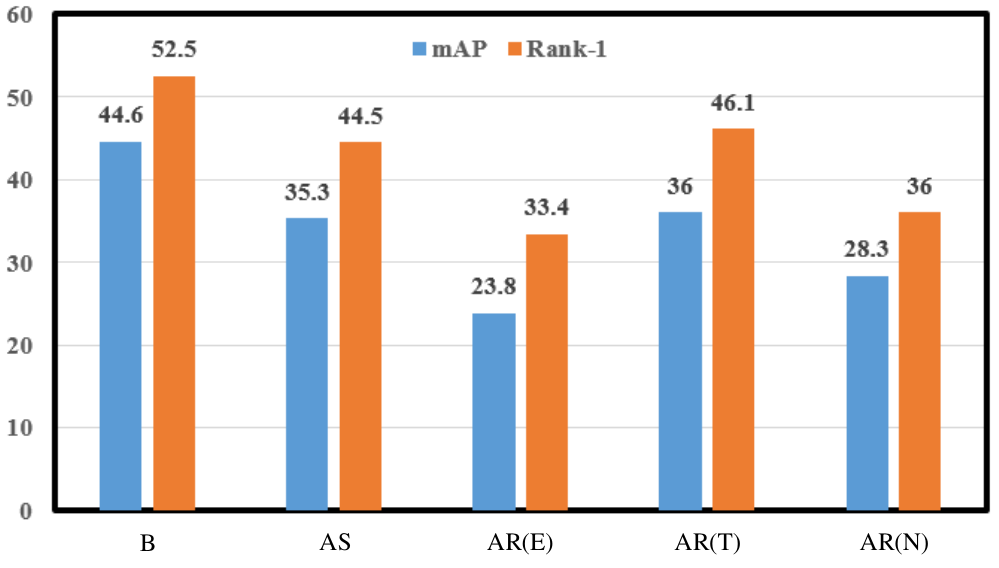}}
\subfigure[] {\includegraphics[height=1.1in,width=1.7in,angle=0]{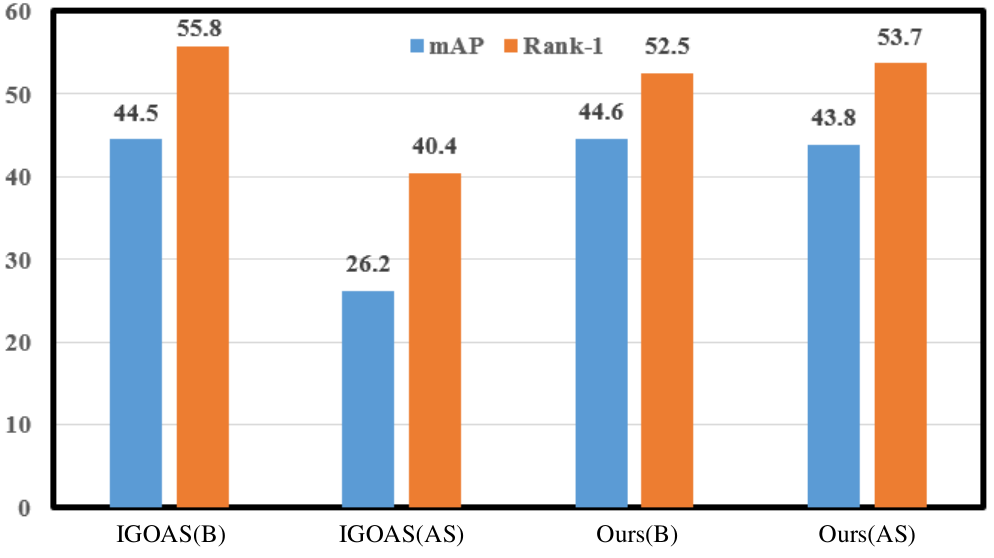}}
\caption{(a) The attack effect of the adversarial example and various adversarial representations on the re-ID model. B: Baseline. AS: Adversarial Samples. AR: Adversarial Representations. E: Erasing. T: Transforming. N: Noising. (b) Comparison of the proposed method and IGOAS in model defense capability.}
\label{Fig:6}
\vspace{-0.3cm}
\end{figure}

\textbf{Model Defensive Capability}. As we discussed above, the model trained by our ETNDNet has a stronger model defensive capability than IGOAS \cite{IGOAS}. In order to further prove this, during testing, we randomly erase the query and gallery samples to further cheat the re-ID system and compare the recognition performance obtained by our method and IGOAS. As illustrated in Figure \ref{Fig:6} (b), the performance that IGOAS achieved was far from the results of its baseline model. Different from it, the proposed ETNDNet gains a recognition rate comparable to the baseline. This is mainly because our adversarial defense strategy enables the re-ID model to pay close attention to as many pedestrian body parts as possible. Even if the information of a certain part is lost, the target pedestrian can still be correctly identified (as illustrated in Figure \ref{Fig:7}). The above qualitative and quantitative analysis fully demonstrate that our method with the developed novel GAN-based adversarial defense is valuable in object re-identification and deserves further exploration.

\begin{figure}[!t]
\centering
\subfigbottomskip=-1pt
\subfigcapskip=-1pt
\subfigure[] {\includegraphics[height=2.8in,width=0.5in,angle=0]{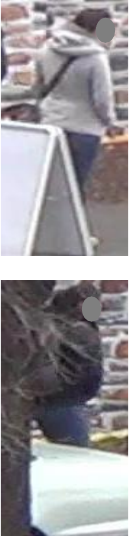}}
\subfigure[] {\includegraphics[height=2.8in,width=0.5in,angle=0]{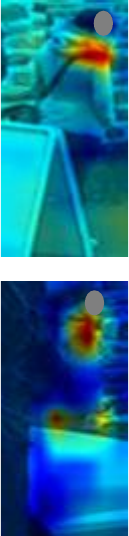}}
\subfigure[] {\includegraphics[height=2.8in,width=0.5in,angle=0]{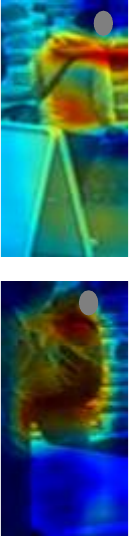}}
\setcounter{subfigure}{0}
\subfigure[] 
{\includegraphics[height=2.8in,width=0.5in,angle=0]{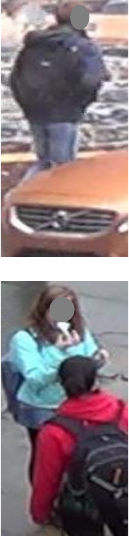}}
\subfigure[] {\includegraphics[height=2.8in,width=0.5in,angle=0]{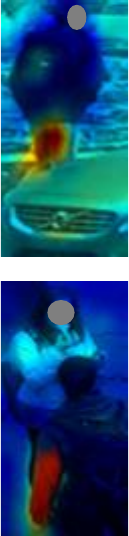}}
\subfigure[] {\includegraphics[height=2.8in,width=0.5in,angle=0]{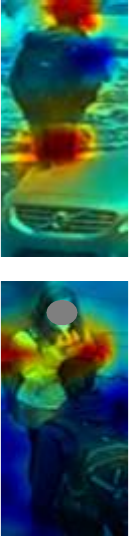}}
\caption{Visualization of regions that the model interest. (a) Original image. (b) Heatmap gained by Baseline. (c) Heatmap gained by ETNDNet.}
\label{Fig:7}
\vspace{-0.3cm}
\end{figure}

\textbf{Limitations}. In the proposed ETNDNet, we perturb feature maps by random erasing, transforming, and noising to form adversarial representations that can simulate various issues caused by occlusions. However, the erased, transformed, and noised areas possibly overlap with occlusions. Additionally, the shapes of obstacles and human bodies are often irregular, while our erased, transformed, and noised areas are regular rectangles, hindering the aggressiveness of adversarial representations and ultimately limiting performance improvement. These limitations prompt us to conduct further investigations in the future.

\section{Conclusion}
In this paper, from the perspective of adversarial defense, we propose a simple yet effective framework named ETNDNet to protect the re-ID system from various occlusion perturbations. Specifically, we generate adversarial representations with incomplete information, misaligned information, and noisy information by erasing, transforming, and noising at feature maps to simulate issues of missing information, position misalignment, and noisy information caused by occlusion. Accordingly, we develop a novel GAN-based adversarial defense approach to train the feature extractor and identity classifier in an adversarial way, thus overcoming the detrimental effects of the aforementioned perturbations on the re-ID system. The proposed ETNDNet exhibits superior results on six public re-ID databases, and extensive experiments fully demonstrate its effectiveness and practicability. In the future, we will further explore the potential of adversarial attack and defense in object re-identification.


%





\ifCLASSOPTIONcaptionsoff
  \newpage
\fi



%
\bibliography{mybibfile}

\begin{thebibliography}{10}
\providecommand{\url}[1]{#1}
\csname url@samestyle\endcsname
\providecommand{\newblock}{\relax}
\providecommand{\bibinfo}[2]{#2}
\providecommand{\BIBentrySTDinterwordspacing}{\spaceskip=0pt\relax}
\providecommand{\BIBentryALTinterwordstretchfactor}{4}
\providecommand{\BIBentryALTinterwordspacing}{\spaceskip=\fontdimen2\font plus
\BIBentryALTinterwordstretchfactor\fontdimen3\font minus \fontdimen4\font\relax}
\providecommand{\BIBforeignlanguage}[2]{{%
\expandafter\ifx\csname l@#1\endcsname\relax
\typeout{** WARNING: IEEEtran.bst: No hyphenation pattern has been}%
\typeout{** loaded for the language `#1'. Using the pattern for}%
\typeout{** the default language instead.}%
\else
\language=\csname l@#1\endcsname
\fi
#2}}
\providecommand{\BIBdecl}{\relax}
\BIBdecl

\bibitem{MCNF}
H.~B. Shitrit, J.~Berclaz, F.~Fleuret, and P.~Fua, ``Multi-commodity network flow for tracking multiple people,'' \emph{IEEE Transactions on Pattern Analysis and Machine Intelligence}, vol.~36, no.~8, pp. 1614--1627, 2014, doi: {10.1109/TPAMI.2013.210}.

\bibitem{DLAD}
Y.~Li, L.~Ma, Z.~Zhong, F.~Liu, M.~A. Chapman, D.~Cao, and J.~Li, ``Deep learning for lidar point clouds in autonomous driving: A review,'' \emph{IEEE Transactions on Neural Networks and Learning Systems}, vol.~32, no.~8, pp. 3412--3432, 2021, doi: {10.1109/TNNLS.2020.3015992}.

\bibitem{CCG-LSTM}
J.~Tang, X.~Shu, R.~Yan, and L.~Zhang, ``Coherence constrained graph lstm for group activity recognition,'' \emph{IEEE Transactions on Pattern Analysis and Machine Intelligence}, vol.~44, no.~2, pp. 636--647, 2022, doi: {10.1109/TPAMI.2019.2928540}.

\bibitem{BAGTRICKS}
H.~Luo, Y.~Gu, X.~Liao, S.~Lai, and W.~Jiang, ``Bag of tricks and a strong baseline for deep person re-identification,'' in \emph{Proceedings of the IEEE Conference on Computer Vision and Pattern Recognition Workshops (CVPRW)}, 2019, pp. 1487--1495, doi: {10.1109/CVPRW.2019.00190}.

\bibitem{AGW}
M.~Ye, J.~Shen, G.~Lin, T.~Xiang, L.~Shao, and S.~C.~H. Hoi, ``Deep learning for person re-identification: A survey and outlook,'' \emph{IEEE Transactions on Pattern Analysis and Machine Intelligence}, vol.~44, no.~6, pp. 2872--2893, 2022, doi: {10.1109/TPAMI.2021.3054775}.

\bibitem{LSST}
X.~Shu, X.~Wang, X.~Zang, S.~Zhang, Y.~Chen, G.~Li, and Q.~Tian, ``Large-scale spatio-temporal person re-identification: Algorithms and benchmark,'' \emph{IEEE Transactions on Circuits and Systems for Video Technology}, vol.~32, no.~7, pp. 4390--4403, 2022, doi: {10.1109/TCSVT.2021.3128214}.

\bibitem{AADIFL}
H.~Li, Y.~Chen, D.~Tao, Z.~Yu, and G.~Qi, ``Attribute-aligned domain-invariant feature learning for unsupervised domain adaptation person re-identification,'' \emph{IEEE Transactions on Information Forensics and Security}, vol.~16, pp. 1480--1494, 2021, doi: {10.1109/TIFS.2020.3036800}.

\bibitem{BPDA}
Y.~Wang, G.~Qi, S.~Li, Y.~Chai, and H.~Li, ``Body part-level domain alignment for domain-adaptive person re-identification with transformer framework,'' \emph{IEEE Transactions on Information Forensics and Security}, vol.~17, pp. 3321--3334, 2022, doi: {10.1109/TIFS.2022.3207893}.

\bibitem{Survey}
Y.~Peng, S.~Hou, C.~Cao, X.~Liu, Y.~Huang, and Z.~He, ``Deep learning-based occluded person re-identification: A survey,'' \emph{arXiv preprint arXiv:2207.14452}, 2022.

\bibitem{PVPM}
S.~Gao, J.~Wang, H.~Lu, and Z.~Liu, ``Pose-guided visible part matching for occluded person reid,'' in \emph{Proceedings of the IEEE Conference on Computer Vision and Pattern Recognition (CVPR)}, 2020, pp. 11\,744--11\,752, doi: {10.1109/CVPR42600.2020.01176}.

\bibitem{HOReID}
G.~Wang, S.~Yang, H.~Liu, Z.~Wang, Y.~Yang, S.~Wang, G.~Yu, E.~Zhou, and J.~Sun, ``High-order information matters: Learning relation and topology for occluded person re-identification,'' in \emph{Proceedings of the IEEE Conference on Computer Vision and Pattern Recognition (CVPR)}, 2020, pp. 6449--6458, doi: {10.1109/CVPR42600.2020.00648}.

\bibitem{RFC}
R.~Hou, B.~Ma, H.~Chang, X.~Gu, S.~Shan, and X.~Chen, ``Feature completion for occluded person re-identification,'' \emph{IEEE Transactions on Pattern Analysis and Machine Intelligence}, vol.~44, no.~9, pp. 4894--4912, 2022, doi: {10.1109/TPAMI.2021.3079910}.

\bibitem{Openpose}
Z.~Cao, T.~Simon, S.-E. Wei, and Y.~Sheikh, ``Realtime multi-person 2d pose estimation using part affinity fields,'' in \emph{Proceedings of the IEEE Conference on Computer Vision and Pattern Recognition (CVPR)}, 2017, pp. 7291--7299, doi: {10.1109/CVPR.2017.143}.

\bibitem{LIP}
X.~Liang, K.~Gong, X.~Shen, and L.~Lin, ``Look into person: Joint body parsing \& pose estimation network and a new benchmark,'' \emph{IEEE Transactions on Pattern Analysis and Machine Intelligence}, vol.~41, no.~4, pp. 871--885, 2018, doi: {10.1109/TPAMI.2018.2820063}.

\bibitem{Explaining}
I.~J. Goodfellow, J.~Shlens, and C.~Szegedy, ``Explaining and harnessing adversarial examples,'' \emph{arXiv preprint arXiv:1412.6572}, 2014.

\bibitem{Stability}
S.~Zheng, Y.~Song, T.~Leung, and I.~Goodfellow, ``Improving the robustness of deep neural networks via stability training,'' in \emph{Proceedings of the IEEE Conference on Computer Vision and Pattern Recognition (CVPR)}, 2016, pp. 4480--4488, doi: {10.1109/CVPR.2016.485}.

\bibitem{Gradient}
C.~Lyu, K.~Huang, and H.-N. Liang, ``A unified gradient regularization family for adversarial examples,'' in \emph{Proceedings of the IEEE International Conference on Data Mining}, 2015, pp. 301--309, doi: {10.1109/ICDM.2015.84}.

\bibitem{AdverOccluded}
H.~Huang, D.~Li, Z.~Zhang, X.~Chen, and K.~Huang, ``Adversarially occluded samples for person re-identification,'' in \emph{Proceedings of the IEEE Conference on Computer Vision and Pattern Recognition (CVPR)}, 2018, pp. 5098--5107, doi: {10.1109/CVPR.2018.00535}.

\bibitem{IGOAS}
C.~Zhao, X.~Lv, S.~Dou, S.~Zhang, J.~Wu, and L.~Wang, ``Incremental generative occlusion adversarial suppression network for person reid,'' \emph{IEEE Transactions on Image Processing}, vol.~30, pp. 4212--4224, 2021, doi: {10.1109/TIP.2021.3070182}.

\bibitem{Bugs}
A.~Ilyas, S.~Santurkar, D.~Tsipras, L.~Engstrom, B.~Tran, and A.~Madry, ``Adversarial examples are not bugs, they are features,'' \emph{Advances in Neural Information Processing Systems}, vol.~32, 2019.

\bibitem{GAN}
I.~Goodfellow, J.~Pouget-Abadie, M.~Mirza, B.~Xu, D.~Warde-Farley, S.~Ozair, A.~Courville, and Y.~Bengio, ``Generative adversarial networks,'' \emph{Communications of the ACM}, vol.~63, no.~11, pp. 139--144, 2020, doi: \href{https://doi.org/10.1145/3422622}{10.1145/3422622}.

\bibitem{GAT}
H.~Lee, S.~Han, and J.~Lee, ``Generative adversarial trainer: Defense to adversarial perturbations with gan,'' \emph{arXiv preprint arXiv:1705.03387}, 2017.

\bibitem{SALF}
M.~Farenzena, L.~Bazzani, A.~Perina, V.~Murino, and M.~Cristani, ``Person re-identification by symmetry-driven accumulation of local features,'' in \emph{Proceedings of the IEEE Conference on Computer Vision and Pattern Recognition (CVPR)}, 2010, pp. 2360--2367, doi: {10.1109/CVPR.2010.5539926}.

\bibitem{LOMO}
S.~Liao, Y.~Hu, X.~Zhu, and S.~Z. Li, ``Person re-identification by local maximal occurrence representation and metric learning,'' in \emph{Proceedings of the IEEE Conference on Computer Vision and Pattern Recognition (CVPR)}, 2015, pp. 2197--2206, doi: {10.1109/CVPR.2015.7298832}.

\bibitem{AIESS}
H.~Li, S.~Yan, Z.~Yu, and D.~Tao, ``Attribute-identity embedding and self-supervised learning for scalable person re-identification,'' \emph{IEEE Transactions on Circuits and Systems for Video Technology}, vol.~30, no.~10, pp. 3472--3485, 2019, doi: {10.1109/TCSVT.2019.2952550}.

\bibitem{DML}
D.~Yi, Z.~Lei, S.~Liao, and S.~Z. Li, ``Deep metric learning for person re-identification,'' in \emph{Proceedings of the IEEE International Conference on Pattern Recognition (ICPR)}, 2014, pp. 34--39, doi: {10.1109/ICPR.2014.16}.

\bibitem{PCB}
Y.~Sun, L.~Zheng, Y.~Yang, Q.~Tian, and S.~Wang, ``Beyond part models: Person retrieval with refined part pooling (and a strong convolutional baseline),'' in \emph{Proceedings of the European Conference on Computer Vision (ECCV)}, 2018, pp. 480--496, doi: \href{https://doi.org/10.1007/978-3-030-01225-0_30}{10.1007/978-3-030-01225-0\_30}.

\bibitem{WSS}
Z.~Jin, J.~Xie, B.~Wu, and L.~Shen, ``Weakly supervised pedestrian segmentation for person re-identification,'' \emph{IEEE Transactions on Circuits and Systems for Video Technology}, vol.~33, no.~3, pp. 1349--1362, 2023, doi: {10.1109/TCSVT.2022.3210476}.

\bibitem{MPLMV}
S.~Li, F.~Li, K.~Wang, G.~Qi, and H.~Li, ``Mutual prediction learning and mixed viewpoints for unsupervised-domain adaptation person re-identification on blockchain,'' \emph{Simulation Modelling Practice and Theory}, vol. 119, p. 102568, 2022, doi: \href{https://doi.org/10.1016/j.simpat.2022.102568}{10.1016/j.simpat.2022.102568}.

\bibitem{SCFC}
Y.~Wang, K.~Xu, Y.~Chai, Y.~Jiang, and G.~Qi, ``Semantic consistent feature construction and multi-granularity feature learning for visible-infrared person re-identification,'' \emph{The Visual Computer}, pp. 1--17, 2023, doi: \href{https://doi.org/10.1007/s00371-023-02923-w}{10.1007/s00371-023-02923-w}.

\bibitem{TALMVR}
H.~Li, N.~Dong, Z.~Yu, D.~Tao, and G.~Qi, ``Triple adversarial learning and multi-view imaginative reasoning for unsupervised domain adaptation person re-identification,'' \emph{IEEE Transactions on Circuits and Systems for Video Technology}, vol.~32, no.~5, pp. 2814--2830, 2022, doi: {10.1109/TCSVT.2021.3099943}.

\bibitem{co-saliency}
J.~Zhuo, J.~Lai, and P.~Chen, ``A novel teacher-student learning framework for occluded person re-identification,'' \emph{arXiv preprint arXiv:1907.03253}, 2019.

\bibitem{HG}
M.~Kiran, R.~G. Praveen, L.~T. Nguyen-Meidine, S.~Belharbi, L.-A. Blais-Morin, and E.~Granger, ``Holistic guidance for occluded person re-identification,'' \emph{arXiv preprint arXiv:2104.06524}, 2021.

\bibitem{FPR}
L.~He, Y.~Wang, W.~Liu, H.~Zhao, Z.~Sun, and J.~Feng, ``Foreground-aware pyramid reconstruction for alignment-free occluded person re-identification,'' in \emph{Proceedings of the IEEE International Conference on Computer Vision (ICCV)}, 2019, pp. 8450--8459, doi: {10.1109/ICCV.2019.00854}.

\bibitem{PGFA}
J.~Miao, Y.~Wu, P.~Liu, Y.~Ding, and Y.~Yang, ``Pose-guided feature alignment for occluded person re-identification,'' in \emph{Proceedings of the IEEE International Conference on Computer Vision (ICCV)}, 2019, pp. 542--551, doi: {10.1109/ICCV.2019.00063}.

\bibitem{SORN}
X.~Zhang, Y.~Yan, J.-H. Xue, Y.~Hua, and H.~Wang, ``Semantic-aware occlusion-robust network for occluded person re-identification,'' \emph{IEEE Transactions on Circuits and Systems for Video Technology}, vol.~31, no.~7, pp. 2764--2778, 2020, doi: {10.1109/TCSVT.2020.3033165}.

\bibitem{PGFL-KD}
K.~Zheng, C.~Lan, W.~Zeng, J.~Liu, Z.~Zhang, and Z.-J. Zha, ``Pose-guided feature learning with knowledge distillation for occluded person re-identification,'' in \emph{Proceedings of the 29th ACM International Conference on Multimedia}, 2021, pp. 4537--4545, doi: \href{https://doi.org/10.1145/3474085.3475610}{10.1145/3474085.3475610}.

\bibitem{BlockMix}
H.~Tang, Z.~Li, Z.~Peng, and J.~Tang, ``Blockmix: meta regularization and self-calibrated inference for metric-based meta-learning,'' in \emph{Proceedings of the 28th ACM International Conference on Multimedia}, 2020, pp. 610--618, doi: \href{https://doi.org/10.1145/3394171.3413884}{10.1145/3394171.3413884}.

\bibitem{Fewshot}
H.~Tang, C.~Yuan, Z.~Li, and J.~Tang, ``Learning attention-guided pyramidal features for few-shot fine-grained recognition,'' \emph{Pattern Recognition}, vol. 130, p. 108792, 2022, doi: \href{https://doi.org/10.1016/j.patcog.2022.108792}{10.1016/j.patcog.2022.108792}.

\bibitem{KGS}
Z.~Li, H.~Tang, Z.~Peng, G.-J. Qi, and J.~Tang, ``Knowledge-guided semantic transfer network for few-shot image recognition,'' \emph{IEEE Transactions on Neural Networks and Learning Systems}, 2023, doi: {10.1109/TNNLS.2023.3240195}.

\bibitem{Boosting}
Z.~Zha, H.~Tang, Y.~Sun, and J.~Tang, ``Boosting few-shot fine-grained recognition with background suppression and foreground alignment,'' \emph{IEEE Transactions on Circuits and Systems for Video Technology}, 2023, doi: {10.1109/TCSVT.2023.3236636}.

\bibitem{Intriguing}
C.~Szegedy, W.~Zaremba, I.~Sutskever, J.~Bruna, D.~Erhan, I.~Goodfellow, and R.~Fergus, ``Intriguing properties of neural networks,'' \emph{arXiv preprint arXiv:1312.6199}, 2013.

\bibitem{Adversarial}
A.~Kurakin, I.~Goodfellow, and S.~Bengio, ``Adversarial machine learning at scale,'' \emph{arXiv preprint arXiv:1611.01236}, 2016.

\bibitem{Advpattern}
Z.~Wang, S.~Zheng, M.~Song, Q.~Wang, A.~Rahimpour, and H.~Qi, ``advpattern: physical-world attacks on deep person re-identification via adversarially transformable patterns,'' in \emph{Proceedings of the IEEE International Conference on Computer Vision (ICCV)}, 2019, pp. 8341--8350, doi: {10.1109/ICCV.2019.00843}.

\bibitem{Transferable}
H.~Wang, G.~Wang, Y.~Li, D.~Zhang, and L.~Lin, ``Transferable, controllable, and inconspicuous adversarial attacks on person re-identification with deep mis-ranking,'' in \emph{Proceedings of the IEEE Conference on Computer Vision and Pattern Recognition (CVPR)}, 2020, pp. 342--351, doi: {10.1109/CVPR42600.2020.00042}.

\bibitem{DeepCloak}
J.~Gao, B.~Wang, Z.~Lin, W.~Xu, and Y.~Qi, ``Deepcloak: Masking deep neural network models for robustness against adversarial samples,'' \emph{arXiv preprint arXiv:1702.06763}, 2017.

\bibitem{Multiexpert}
X.~Wang, S.~Li, M.~Liu, Y.~Wang, and A.~K. Roy-Chowdhury, ``Multi-expert adversarial attack detection in person re-identification using context inconsistency,'' in \emph{Proceedings of the IEEE International Conference on Computer Vision (ICCV)}, 2021, pp. 15\,097--15\,107, doi: {10.1109/ICCV48922.2021.01482}.

\bibitem{Resnet}
K.~He, X.~Zhang, S.~Ren, and J.~Sun, ``Deep residual learning for image recognition,'' in \emph{Proceedings of the IEEE Conference on Computer Vision and Pattern Recognition (CVPR)}, 2016, pp. 770--778, doi: {10.1109/CVPR.2016.90}.

\bibitem{Imagenet}
J.~Deng, W.~Dong, R.~Socher, L.-J. Li, K.~Li, and L.~Fei-Fei, ``Imagenet: A large-scale hierarchical image database,'' in \emph{Proceedings of the IEEE Conference on Computer Vision and Pattern Recognition (CVPR)}, 2009, pp. 248--255, doi: {10.1109/CVPR.2009.5206848}.

\bibitem{LS}
C.~Szegedy, V.~Vanhoucke, S.~Ioffe, J.~Shlens, and Z.~Wojna, ``Rethinking the inception architecture for computer vision,'' in \emph{Proceedings of the IEEE Conference on Computer Vision and Pattern Recognition (CVPR)}, 2016, pp. 2818--2826, doi: {10.1109/CVPR.2016.308}.

\bibitem{RE}
Z.~Zhong, L.~Zheng, G.~Kang, S.~Li, and Y.~Yang, ``Random erasing data augmentation,'' in \emph{Proceedings of the AAAI Conference on Artificial Intelligence}, vol.~34, no.~07, 2020, pp. 13\,001--13\,008, doi: \href{https://doi.org/10.1609/aaai.v34i07.7000}{10.1609/aaai.v34i07.7000}.

\bibitem{Duke}
E.~Ristani, F.~Solera, R.~Zou, R.~Cucchiara, and C.~Tomasi, ``Performance measures and a data set for multi-target, multi-camera tracking,'' in \emph{Proceedings of the European Conference on Computer Vision (ECCV)}.\hskip 1em plus 0.5em minus 0.4em\relax Springer, 2016, pp. 17--35, doi: \href{https://doi.org/10.1007/978-3-319-48881-3_2}{10.1007/978-3-319-48881-3\_2}.

\bibitem{OccludedReID}
J.~Zhuo, Z.~Chen, J.~Lai, and G.~Wang, ``Occluded person re-identification,'' in \emph{IEEE International Conference on Multimedia and Expo (ICME)}, 2018, pp. 1--6, doi: {10.1109/ICME.2018.8486568}.

\bibitem{Market}
L.~Zheng, L.~Shen, L.~Tian, S.~Wang, J.~Wang, and Q.~Tian, ``Scalable person re-identification: A benchmark,'' in \emph{Proceedings of the IEEE International Conference on Computer Vision (ICCV)}, 2015, pp. 1116--1124, doi: {10.1109/ICCV.2015.133}.

\bibitem{PTGAN}
L.~Wei, S.~Zhang, W.~Gao, and Q.~Tian, ``Person transfer gan to bridge domain gap for person re-identification,'' in \emph{Proceedings of the IEEE Conference on Computer Vision and Pattern Recognition (CVPR)}, 2018, pp. 79--88, doi: {10.1109/CVPR.2018.00016}.

\bibitem{GASM}
L.~He and W.~Liu, ``Guided saliency feature learning for person re-identification in crowded scenes,'' in \emph{Proceedings of the European Conference on Computer Vision (ECCV)}.\hskip 1em plus 0.5em minus 0.4em\relax Springer, 2020, pp. 357--373, doi: \href{https://doi.org/10.1007/978-3-030-58604-1_22}{10.1007/978-3-030-58604-1\_22}.

\bibitem{OAMN}
P.~Chen, W.~Liu, P.~Dai, J.~Liu, Q.~Ye, M.~Xu, Q.~Chen, and R.~Ji, ``Occlude them all: Occlusion-aware attention network for occluded person re-id,'' in \emph{2021 IEEE/CVF International Conference on Computer Vision (ICCV)}, 2021, pp. 11\,813--11\,822, doi: {10.1109/ICCV48922.2021.01162}.

\bibitem{PIRT}
Z.~Ma, Y.~Zhao, and J.~Li, ``Pose-guided inter-and intra-part relational transformer for occluded person re-identification,'' in \emph{Proceedings of the 29th ACM International Conference on Multimedia}, 2021, pp. 1487--1496, doi: \href{https://doi.org/10.1145/3474085.3475283}{10.1145/3474085.3475283}.

\bibitem{PGMANet}
Y.~Zhai, X.~Han, W.~Ma, X.~Gou, and G.~Xiao, ``Pgmanet: Pose-guided mixed attention network for occluded person re-identification,'' in \emph{2021 International Joint Conference on Neural Networks (IJCNN)}.\hskip 1em plus 0.5em minus 0.4em\relax IEEE, 2021, pp. 1--8, doi: {10.1109/IJCNN52387.2021.9534442}.

\bibitem{PEFB}
J.~Miao, Y.~Wu, and Y.~Yang, ``Identifying visible parts via pose estimation for occluded person re-identification,'' \emph{IEEE Transactions on Neural Networks and Learning Systems}, 2021, doi: {10.1109/TNNLS.2021.3059515}.

\bibitem{ISP}
K.~Zhu, H.~Guo, Z.~Liu, M.~Tang, and J.~Wang, ``Identity-guided human semantic parsing for person re-identification,'' in \emph{Proceedings of the European Conference on Computer Vision (ECCV)}.\hskip 1em plus 0.5em minus 0.4em\relax Springer, 2020, pp. 346--363, doi: \href{https://doi.org/10.1007/978-3-030-58580-8_21}{10.1007/978-3-030-58580-8\_21}.

\bibitem{ASAN}
H.~Jin, S.~Lai, and X.~Qian, ``Occlusion-sensitive person re-identification via attribute-based shift attention,'' \emph{IEEE Transactions on Circuits and Systems for Video Technology}, vol.~32, no.~4, pp. 2170--2185, 2021, doi: {10.1109/TCSVT.2021.3088446}.

\bibitem{CBDBNet}
H.~Tan, X.~Liu, Y.~Bian, H.~Wang, and B.~Yin, ``Incomplete descriptor mining with elastic loss for person re-identification,'' \emph{IEEE Transactions on Circuits and Systems for Video Technology}, vol.~32, no.~1, pp. 160--171, 2021, doi: {10.1109/TCSVT.2021.3061412}.

\bibitem{FED}
Z.~Wang, F.~Zhu, S.~Tang, R.~Zhao, L.~He, and J.~Song, ``Feature erasing and diffusion network for occluded person re-identification,'' in \emph{Proceedings of the IEEE/CVF Conference on Computer Vision and Pattern Recognition}, 2022, pp. 4754--4763, doi: {10.1109/CVPR52688.2022.00471}.

\bibitem{QPM}
P.~Wang, C.~Ding, Z.~Shao, Z.~Hong, S.~Zhang, and D.~Tao, ``Quality-aware part models for occluded person re-identification,'' \emph{IEEE Transactions on Multimedia}, 2022, doi: {10.1109/TMM.2022.3156282}.

\bibitem{DRLNet}
M.~Jia, X.~Cheng, S.~Lu, and J.~Zhang, ``Learning disentangled representation implicitly via transformer for occluded person re-identification,'' \emph{IEEE Transactions on Multimedia}, 2022, doi: {10.1109/TMM.2022.3141267}.

\bibitem{OPR-DAAO}
S.~Wang, R.~Liu, H.~Li, G.~Qi, and Z.~Yu, ``Occluded person re-identification via defending against attacks from obstacles,'' \emph{IEEE Transactions on Information Forensics and Security}, vol.~18, pp. 147--161, 2022, doi: {10.1109/TIFS.2022.3218449}.

\bibitem{RTGAT}
M.~Huang, C.~Hou, Q.~Yang, and Z.~Wang, ``Reasoning and tuning: Graph attention network for occluded person re-identification,'' \emph{IEEE Transactions on Image Processing}, vol.~32, pp. 1568--1582, 2023, doi: {10.1109/TIP.2023.3247159}.

\bibitem{Adam}
D.~P. Kingma and J.~Ba, ``Adam: A method for stochastic optimization,'' \emph{arXiv preprint arXiv:1412.6980}, 2014.

\bibitem{IDE}
L.~Zheng, H.~Zhang, S.~Sun, M.~Chandraker, Y.~Yang, and Q.~Tian, ``Person re-identification in the wild,'' in \emph{Proceedings of the IEEE Conference on Computer Vision and Pattern Recognition (CVPR)}, 2017, pp. 1367--1376, doi: {10.1109/CVPR.2017.357}.

\bibitem{Non-Local}
X.~Wang, R.~Girshick, A.~Gupta, and K.~He, ``Non-local neural networks,'' in \emph{Proceedings of the IEEE conference on computer vision and pattern recognition}, 2018, pp. 7794--7803, doi: {10.1109/CVPR.2018.00813}.

\bibitem{VPM}
Y.~Sun, Q.~Xu, Y.~Li, C.~Zhang, Y.~Li, S.~Wang, and J.~Sun, ``Perceive where to focus: Learning visibility-aware part-level features for partial person re-identification,'' in \emph{Proceedings of the IEEE Conference on Computer Vision and Pattern Recognition (CVPR)}, 2019, pp. 393--402, doi: {10.1109/CVPR.2019.00048}.

\bibitem{IANet}
R.~Hou, B.~Ma, H.~Chang, X.~Gu, S.~Shan, and X.~Chen, ``Interaction-and-aggregation network for person re-identification,'' in \emph{Proceedings of the IEEE Conference on Computer Vision and Pattern Recognition (CVPR)}, 2019, pp. 9317--9326, doi: {10.1109/CVPR.2019.00954}.

\bibitem{MHSANet}
H.~Tan, X.~Liu, B.~Yin, and X.~Li, ``Mhsa-net: Multihead self-attention network for occluded person re-identification,'' \emph{IEEE Transactions on Neural Networks and Learning Systems}, 2022, doi: {10.1109/TNNLS.2022.3144163}.

\bibitem{OSNet}
K.~Zhou, Y.~Yang, A.~Cavallaro, and T.~Xiang, ``Omni-scale feature learning for person re-identification,'' in \emph{Proceedings of the IEEE Conference on Computer Vision and Pattern Recognition (CVPR)}, 2019, pp. 3702--3712, doi: {10.1109/ICCV.2019.00380}.

\bibitem{POS}
S.~Wang, B.~Huang, H.~Li, G.~Qi, D.~Tao, and Z.~Yu, ``Key point-aware occlusion suppression and semantic alignment for occluded person re-identification,'' \emph{Information Sciences}, vol. 606, pp. 669--687, 2022, doi: \href{https://doi.org/10.1016/j.ins.2022.05.077}{10.1016/j.ins.2022.05.077}.

\bibitem{PLIP}
J.~Zuo, C.~Yu, N.~Sang, and C.~Gao, ``Plip: Language-image pre-training for person representation learning,'' \emph{arXiv preprint arXiv:2305.08386}, 2023.

\bibitem{BPBreID}
V.~Somers, C.~De~Vleeschouwer, and A.~Alahi, ``Body part-based representation learning for occluded person re-identification,'' in \emph{Proceedings of the IEEE/CVF Winter Conference on Applications of Computer Vision}, 2023, pp. 1613--1623, doi: {10.1109/WACV56688.2023.00166}.

\bibitem{GPS}
B.~X. Nguyen, B.~D. Nguyen, T.~Do, E.~Tjiputra, Q.~D. Tran, and A.~Nguyen, ``Graph-based person signature for person re-identifications,'' in \emph{Proceedings of the IEEE/CVF conference on computer vision and pattern recognition}, 2021, pp. 3492--3501, doi: {10.1109/CVPRW53098.2021.00388}.

\bibitem{DropBlock}
Z.~Dai, M.~Chen, X.~Gu, S.~Zhu, and P.~Tan, ``Batch dropblock network for person re-identification and beyond,'' in \emph{Proceedings of the IEEE/CVF international conference on computer vision}, 2019, pp. 3691--3701, doi: {10.1109/ICCV.2019.00379}.

\bibitem{SNR}
X.~Jin, C.~Lan, W.~Zeng, Z.~Chen, and L.~Zhang, ``Style normalization and restitution for generalizable person re-identification,'' in \emph{proceedings of the IEEE/CVF conference on computer vision and pattern recognition}, 2020, pp. 3143--3152, doi: {10.1109/CVPR42600.2020.00321}.

\bibitem{QAConv}
S.~Liao and L.~Shao, ``Interpretable and generalizable person re-identification with query-adaptive convolution and temporal lifting,'' in \emph{Computer Vision--ECCV 2020: 16th European Conference, Glasgow, UK, August 23--28, 2020, Proceedings, Part XI 16}.\hskip 1em plus 0.5em minus 0.4em\relax Springer, 2020, pp. 456--474, doi: \href{https://doi.org/10.1007/978-3-030-58621-8_27}{10.1007/978-3-030-58621-8\_27}.

\end{thebibliography}


\end{document}